\documentclass[10pt,twocolumn]{article}

\usepackage{times}
\usepackage{epsfig}
\usepackage{amssymb}
\usepackage{amsmath}
\usepackage{lineno}
\usepackage{authblk}
\usepackage{hyperref}
\usepackage{multirow}
\usepackage{booktabs}
\usepackage{xcolor}
\usepackage{colortbl}
\usepackage{adjustbox}
\usepackage{graphicx}
\usepackage{subcaption}
\usepackage{pifont}
\usepackage[T1]{fontenc}

\newcommand{\cmark}{\ding{51}}
\newcommand{\xmark}{\textcolor{black!35}{\ding{55}}}

\hypersetup{
  colorlinks=true,
  linkcolor=blue,
  citecolor=blue,
  urlcolor=blue,
  pdfborder={0 0 0},
}

\hypersetup{
    pdftitle={Are Textual Residuals Truly Complementary? Learning Visual Complements under Semantic Superposition for Visible-Infrared Person Re-Identification},
    pdfauthor={Xiaoyang Li, Mingming Lu, Wei Zhou, Hao Li, Canhui Zhang}
}

\title{Are Textual Residuals Truly Complementary? Learning Visual Complements under Semantic Superposition for Visible-Infrared Person Re-Identification}

\author[1]{Xiaoyang Li}
\author[1]{Mingming Lu\thanks{Corresponding author. Email: mingminglu@csu.edu.cn}}
\author[1]{Wei Zhou}
\author[1]{Hao Li}
\author[1]{Canhui Zhang}

\affil[1]{School of Computer Science and Engineering, Central South University, Changsha 410083, Hunan Province, China}

\date{}

\begin{document}
\maketitle

\begin{abstract}
Text-enhanced visible-infrared person re-identification (VI-ReID) exploits textual descriptions to compensate for appearance cues that are weak or unavailable under infrared imaging. Existing methods construct compensation from differences between modality-oriented textual representations, implicitly assuming that textual differences are visually complementary. We revisit this assumption through semantic superposition and show that dense textual residuals may entangle useful appearance variations with redundant, making direct residual fusion unreliable. We propose Superposition-Aware Residual Calibration, which transforms dense textual differences into selective and visually grounded complements. Low-Interference Sparse Residual Learning represents modality-oriented text embeddings as sparse activation strengths over a shared dictionary, selects positive strength differences, and suppresses both dictionary-level overlap and activation-weighted interference to reconstruct a compact, selectively controlled residual. Visual Complement and Counterfactual Calibration further calibrates the residual against identity-specific RGB-IR visual discrepancies, encouraging retained semantics to provide meaningful compensation while preserving the residual structure. Experiments on Tri-SYSU-MM01, Tri-RegDB, and Tri-LLCM show that SARC consistently improves the VI-ReID backbone and remains competitive with text-assisted methods. Mechanism analyses show that SARC produces more concentrated textual variations, reduces dictionary-level feature interference, and improves residual-visual-complement alignment for most samples. These results demonstrate that effective cross-modal compensation requires textual residuals to be selectively represented and explicitly calibrated, rather than directly treated as visual complements. Code and implementation details will be released upon publication.
\end{abstract}

\begin{figure*}[t]
    \centering
    \includegraphics[width=\textwidth]{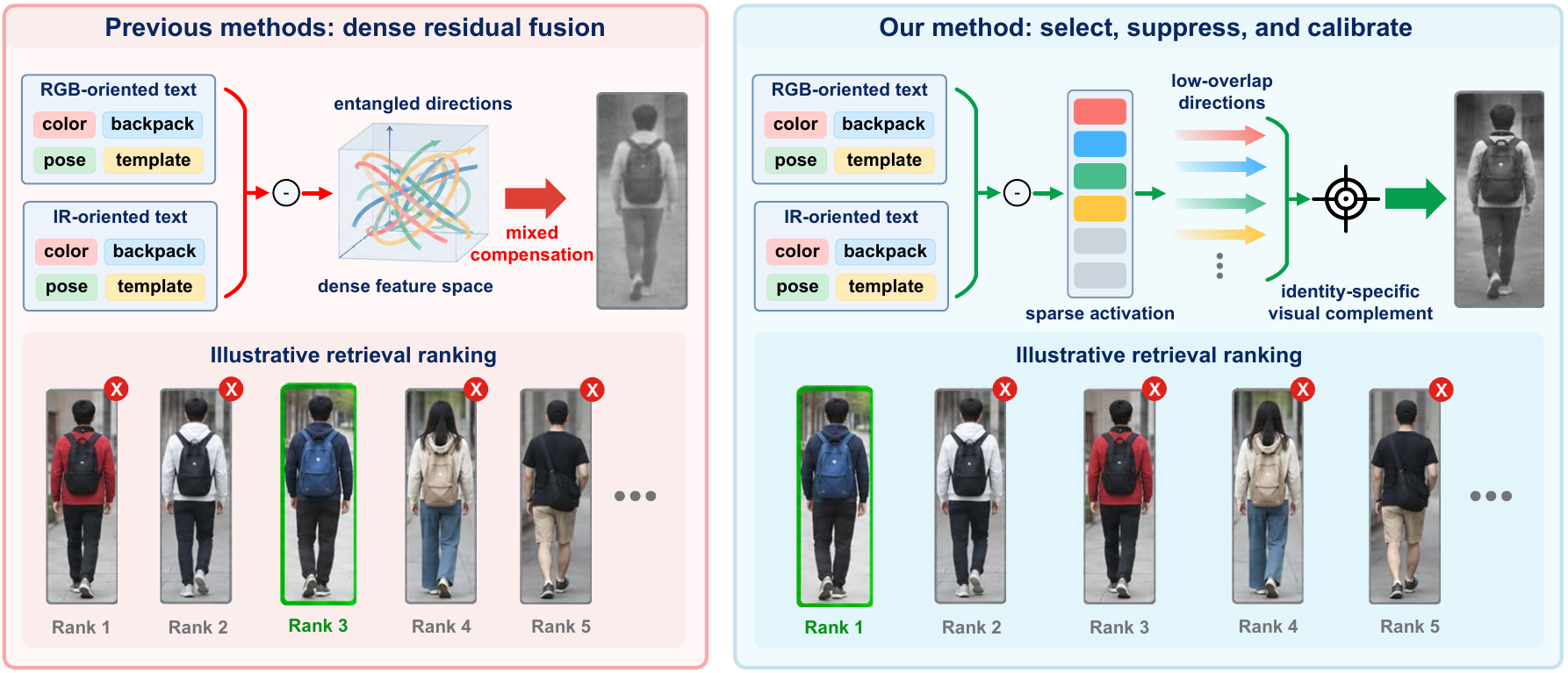}
    \caption{Conceptual comparison between direct dense residual fusion and the proposed SARC. Existing strategies inject a dense textual residual into the infrared representation, where overlapping semantic factors can produce mixed compensation. SARC first represents textual changes as selectable sparse feature strengths, suppresses interference among overlapping directions, and calibrates the reconstructed residual toward the identity-specific RGB--IR visual complement.}
    \label{fig1}
\end{figure*}

\section{Introduction}
\label{sec1}

Visible-infrared person re-identification (VI-ReID) aims to retrieve the same person across visible and infrared cameras, and is essential for around-the-clock intelligent surveillance. Its central challenge arises from the highly heterogeneous visual evidence captured by the two modalities. Visible images preserve fine-grained appearance cues, including clothing colors, textures, and carried objects, whereas infrared images mainly reveal thermal responses, body contours, and coarse structural patterns. Although existing VI-ReID methods reduce modality discrepancy by learning modality-invariant representations~\cite{wang2019learning,choi2020hi,park2021learning}, alignment alone remains insufficient to recover or exploit discriminative visible appearance cues that become weak or unavailable in infrared observations~\cite{wu2021discover,zhang2022fmcnet}.

Text offers a promising route for compensating for this missing evidence~\cite{hu2024empowering,yu2025clip}. Descriptions generated from visible observations can explicitly express colors, accessories, and other identity-related attributes that are difficult to infer from infrared images alone. This idea is closely related to \emph{multi-modal analogical reasoning}. When image and text features are aligned in a unified embedding space~\cite{radford2021learning}, an attribute-level textual displacement can act as a transferable semantic direction. For example, removing the text direction corresponding to ``blue'' from the representation of a blue car and adding the direction corresponding to ``red'' can move the representation toward that of a red car \cite{hu2024empowering}. In text-enhanced VI-ReID, this observation motivates the use of the difference between an RGB-oriented description and an IR-oriented description as a residual that compensates the infrared visual feature. Intuitively, the residual should describe visible information that is present in RGB observations but missing from their infrared counterparts.

However, the analogy-based intuition does not guarantee that a textual residual is a valid visual complement. Existing residual-fusion strategies commonly construct a dense textual difference and inject it directly into the infrared representation~\cite{hu2024empowering}. As illustrated in Fig.~\ref{fig1}, directly constructing and fusing a dense textual difference treats the textual variation as an inseparable semantic displacement. Due to semantic superposition, however, dense representations may encode multiple semantic factors in overlapping feature directions~\cite{elhage2022toy,huben2024sparse}. Consequently, such a residual may indeed transfer useful appearance attributes, such as color or backpack presence, but can simultaneously contain redundant, linguistically induced, or identity-irrelevant information. Direct dense fusion may therefore yield a mixed compensation that does not faithfully reflect the visual evidence missing from the target infrared observation.

We investigate this issue through the lens of \emph{semantic superposition}. A learned representation may encode many co-occurring semantic factors within a finite-dimensional space, giving rise to superposed or polysemantic feature directions~\cite{elhage2022toy,huben2024sparse}. From this perspective, concepts such as clothing color, backpack, pose, linguistic template, and background context may occupy partially overlapping feature directions rather than being represented independently~\cite{gao2025scaling}. Under semantic superposition, the difference between two text embeddings is therefore not necessarily a clean attribute transformation. Instead, it can become an entangled displacement that combines appearance-relevant changes with correlated but visually unsupported semantics~\cite{dong2024embedding}. Directly injecting such a residual into an infrared feature can disturb its identity-discriminative structure and degrade the ranking of difficult positive samples.

This observation raises the central question of this work: \emph{Are textual residuals truly complementary?} We argue that a useful textual residual should satisfy two requirements. First, it should be expressed through a compact set of semantic variations with limited mutual interference. Second, it should be consistent with the visual discrepancy observed between RGB and infrared samples of the same identity.

To this end, we propose \textbf{SARC}, a \textbf{S}uperposition-\textbf{A}ware \textbf{R}esidual \textbf{C}alibration framework for text-enhanced VI-ReID. SARC represents RGB-oriented and IR-oriented textual features as sparse feature strengths over a shared set of learnable directions. It constructs the residual from selected strength variations rather than directly transferring the entire dense textual difference. SARC further learns low-interference feature directions and discourages harmful co-activation among residual components, thereby making textual changes more selective before they are reconstructed in the visual representation space.

More importantly, SARC does not assume that sparse decomposition itself makes a residual visually complementary. During training, it calibrates the constructed residual against the identity-level RGB--IR visual discrepancy. Cross-identity counterfactual supervision further discourages generic directions that are equally compatible with different identities. Therefore, SARC connects the linguistic motivation of multi-modal analogical reasoning with an explicit visual verification mechanism: it preserves useful attribute-level textual changes while suppressing components that lack identity-specific visual support.

The contributions of this work are summarized as follows:

\begin{itemize}
    \item We revisit text-enhanced VI-ReID from the perspective of semantic superposition and challenge the common assumption that textual differences are inherently visually complementary. We distinguish dense textual variation, sparse semantic variation, and identity-specific visual complementarity.

    \item We propose an interference-aware feature-strength decomposition that represents textual features through sparse strengths over shared learnable directions. By suppressing direction overlap and selecting salient strength variations, SARC constructs more selective textual residuals than direct dense subtraction.

    \item We introduce identity-level visual complement calibration with cross-identity counterfactual supervision, encouraging textual residuals to compensate for identity-relevant RGB--IR discrepancies rather than generic modality differences.

    \item Extensive experiments and mechanism analyses demonstrate the effectiveness of SARC. Beyond retrieval performance, the analyses show that sparse coding concentrates textual changes into compact feature directions, low-interference learning reduces dictionary overlap, and calibration conservatively improves residual--complement alignment.
\end{itemize}

\section{Related Work}
\label{sec2}

\subsection{Visible--Infrared Person Re-Identification}

VI-ReID aims to match identities across visible and infrared images despite substantial modality-induced appearance discrepancies~\cite{wu2017rgb}. Existing methods address these discrepancies through image translation, shared representation learning, and embedding augmentation~\cite{wang2019rgb,liu2020parameter,zhang2023diverse}. More recent approaches explore spatial and frequency-domain representations: DSMCL~\cite{zhao2026dual} promotes modality consistency in both domains, while MFEN~\cite{li2026mfen} models information across multiple frequency bands. These approaches seek to suppress modality-specific variation while preserving identity discrimination. However, reducing the discrepancy between modalities does not necessarily recover appearance cues that are weak or absent in infrared observations, motivating the use of additional semantic information for compensation.

Text offers a complementary source of appearance information, and vision--language pretraining provides a foundation for associating textual semantics with visual representations~\cite{radford2021learning}. Text-assisted VI-ReID methods differ in how they incorporate this information. Some use text to supervise visual representation learning, for example through attribute consistency~\cite{yan2025shapespeak}. DSFAD~\cite{dong2025diverse} and CLIP4VI-ReID~\cite{yang2026clip4vi} likewise exploit textual semantics during learning while retaining visual-only inference. Other methods incorporate text at query time to directly compensate infrared representations~\cite{hu2024empowering}. This distinction is consequential: training-time textual supervision influences the learned visual features, whereas query-time compensation additionally depends on the availability and relevance of the supplied descriptions. The two settings therefore differ in both their inference requirements and the information available for retrieval.

SARC follows the query-time compensation setting of TVI-LFM~\cite{hu2024empowering}, which uses differences between IR- and RGB-oriented descriptions to enrich infrared features through multimodal analogical reasoning. Although these textual differences can encode complementary appearance cues, they may also contain variation that is irrelevant to identity or unsupported by the visual representation. The resulting challenge is to determine which components of the textual difference should contribute to visual compensation. Under the same heterogeneous-text protocol, SARC addresses this challenge through sparse selection and visual calibration of textual residuals. Specifically, it selects residual components for compensation and calibrates their contribution using visual evidence. This formulation shifts the focus from incorporating textual differences to controlling which differences are transferred and how they contribute to identity discrimination.

\subsection{Semantic Superposition and Sparse Representation}

The effectiveness of textual residual fusion depends on whether the representation space organizes semantic factors in a way that allows meaningful variations to be isolated. Under semantic superposition, multiple features can share overlapping directions in a finite-dimensional embedding space, enabling representations to encode diverse concepts while introducing interference between them~\cite{elhage2022toy}. Consequently, the difference between two dense textual representations does not necessarily correspond to a single appearance change. Instead, it may contain a mixture of identity-relevant attributes, contextual variations, and correlated semantic components. For residual-based VI-ReID, the key challenge is therefore not simply extracting textual differences, but selectively identifying variations that can provide reliable visual compensation.

Sparse dictionary representations provide a natural mechanism for separating feature directions from their activation strengths. Recent studies on sparse autoencoders show that dense neural activations can be decomposed into sparse combinations of learned latent directions~\cite{bricken2023towards}. Further developments introduce activation constraints such as Top-$k$ coding \cite{gao2025scaling} and improve the separation between direction learning and activation estimation~\cite{rajamanoharan2024improving}. These studies suggest that sparsity can make individual contributions easier to select, while the geometry of the learned dictionary determines the degree of overlap among candidate directions.

Alternative approaches, such as dimension-preserving transformations of vision--language embeddings, provide another way to reduce semantic entanglement. For example, CEDAR~\cite{kubaty2026conceptualizing} learns an invertible transformation with sparse bottlenecks to expose structured representations without increasing dimensionality. Different from these approaches, SARC does not aim to obtain a universally disentangled embedding space. Instead, it uses a shared encoder and overcomplete dictionary to establish comparable feature-strength coordinates for the two descriptions, and combines sparse selection with low-interference constraints to construct task-specific residuals for cross-modal retrieval. The additional latent directions provide more candidate components for residual construction, while low-interference regularization controls their redundancy.

Therefore, sparse decomposition in SARC serves as an intermediate mechanism rather than the final objective. The goal is not merely to obtain sparse or interpretable representations, but to construct residuals whose activated components correspond to missing visual evidence. Accordingly, SARC combines feature-strength selection with dictionary- and activation-level interference control, and further uses identity-level visual supervision to calibrate the selected residuals toward reliable RGB--IR complementarity.

\subsection{Cross-Modal Alignment and Visual Calibration}

Cross-modal alignment seeks to establish correspondence between modalities while preserving identity-discriminative information. Existing methods impose this correspondence at different levels. Image-level objectives encourage overall compatibility between representations, spatial correspondence addresses local misalignment~\cite{park2021learning}, and identity-center objectives constrain cross-modal identity structure by aggregating observations~\cite{liu2020parameter}. Text--visual alignment extends these objectives to linguistic descriptions, encouraging semantic cues to remain consistent with visual evidence~\cite{hu2024empowering,yan2025shapespeak}. Together, these approaches provide a foundation for learning representations that are comparable across modalities and discriminative across identities.

For textual residual compensation, however, alignment of the resulting representation does not explicitly identify the contribution of the transferred residual. A residual may improve compatibility with visible features without providing additional evidence that distinguishes the target identity from competing identities. Residual compensation therefore calls for supervision at the component level: the transferred information should reflect an identity-relevant visual difference, and its contribution should be more useful for the target identity than an identity-mismatched alternative. These requirements concern both the direction of compensation and its comparative identity utility.

Counterfactual supervision offers a way to examine such contributions by comparing outcomes under controlled interventions. This principle has already been explored in VI-ReID. CIFT~\cite{li2022counterfactual} uses Counterfactual Relation Intervention (CRI) to supervise graph affinity and topology learning, while counterfactual attention alignment~\cite{sun2023counterfactual} evaluates and supervises attention through counterfactual interventions. These studies establish the relevance of counterfactual reasoning to VI-ReID, but apply interventions to relations or attention rather than to the textual residual used for visual compensation.

SARC combines visual-reference supervision with cross-identity residual intervention to calibrate textual compensation. During training, identity-level RGB--IR differences provide visual complement targets for refining textual residuals through a learned adapter. This supervision guides the correction toward identity-specific visual differences while retaining information from the individual descriptions. Complementing this target-alignment objective, SARC replaces the matched textual residual with a cross-identity residual while keeping the infrared query and target identity fixed. The counterfactual objective encourages the matched residual to yield greater compatibility with the target identity than the mismatched alternative. SARC thus applies counterfactual supervision to the identity-specific utility of textual compensation and combines it with explicit visual complement targets. These identity-level references are required only during training; at inference, the learned adapter calibrates textual residuals without access to identity prototypes.

\begin{figure*}[t]
    \centering
    \includegraphics[
        width=\textwidth,
        height=0.8\textheight,
        keepaspectratio
    ]{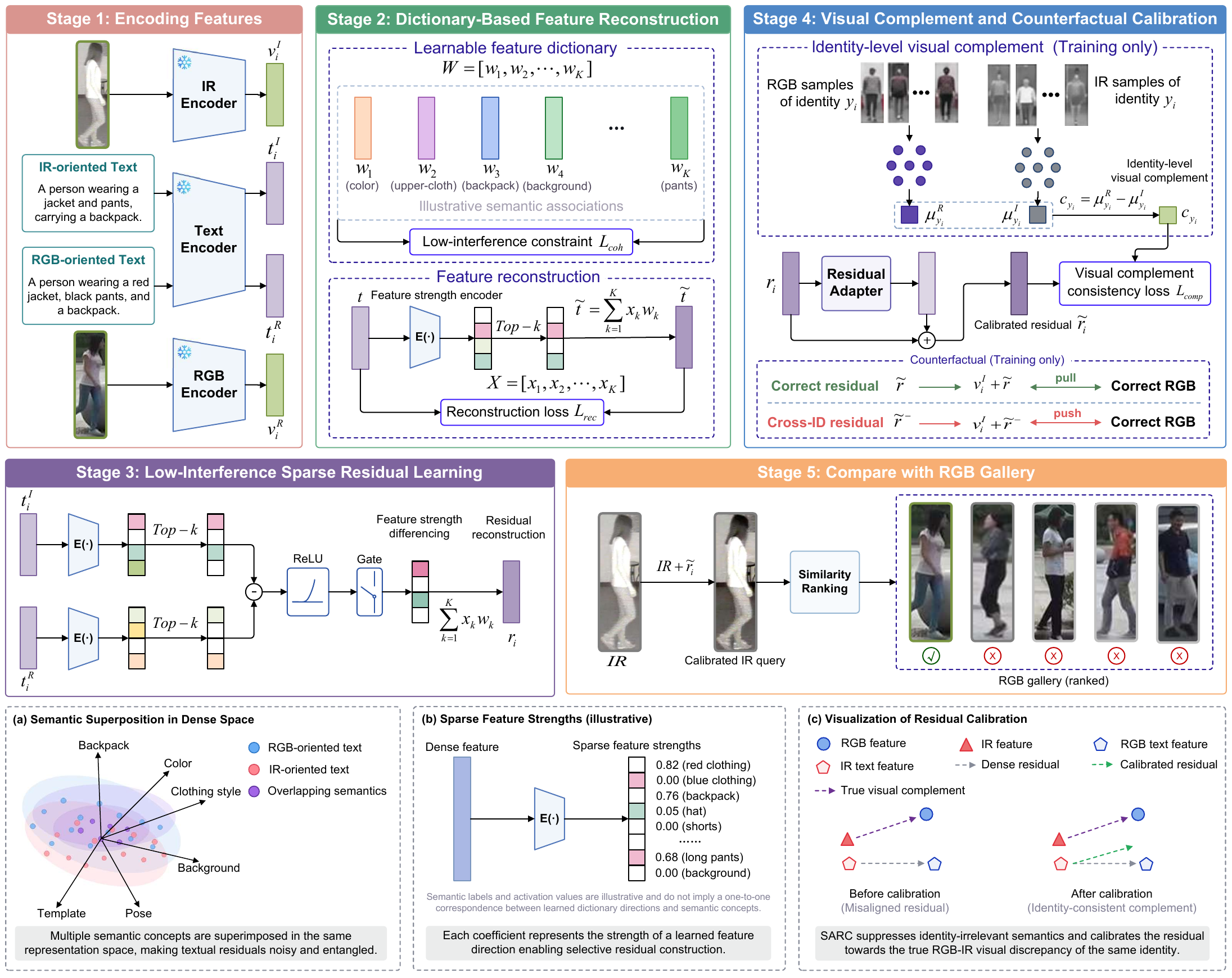}
    \caption{Overview of SARC. Image and text encoders extract multimodal representations, which share a dictionary of learnable feature directions. Low-interference sparse residual learning selects strength variations between RGB- and IR-oriented descriptions and reconstructs a textual residual. Visual complement and counterfactual calibration use identity-level RGB--IR discrepancies and cross-identity residuals as training supervision. The calibrated residual supplements the infrared query representation for retrieval against the RGB gallery. The bottom panels illustrate semantic superposition, sparse feature-strength representation, and visual complement calibration.}
    \label{fig2}
\end{figure*}

\section{Problem Formulation and Motivation}
\label{sec3}

\subsection{Text-Enhanced VI-ReID Setting}
\label{sec3.1}

Textual descriptions provide appearance cues that may be weak or absent in infrared observations. Let $\mathcal{D}=\{\mathcal{D}_{R}^{I},\mathcal{D}_{I}^{I}, \mathcal{D}_{R}^{T},\mathcal{D}_{I}^{T},\mathcal{Y}\}$ denote the dataset, where $\mathcal{D}_{R}^{I}=\{V_i^{R}\}_{i=1}^{N_R}$ and $\mathcal{D}_{I}^{I}=\{V_i^{I}\}_{i=1}^{N_I}$ are the RGB and infrared image sets. Their corresponding descriptions are $\mathcal{D}_{R}^{T}=\{T_i^{R}\}_{i=1}^{N_R}$ and $\mathcal{D}_{I}^{T}=\{T_i^{I}\}_{i=1}^{N_I}$, and $\mathcal{Y}=\{y_i\}_{i=1}^{N_{\mathrm{id}}}$ contains the distinct identity labels.

We follow the heterogeneous-text protocol of TVI-LFM~\cite{hu2024empowering}, which provides an identity-associated RGB-oriented description as auxiliary query-side metadata. Each query is $\mathcal{Q}_i=\{V_i^{I},T_i^{I},T_i^{R}\}$, where $T_i^{R}$ here denotes a description randomly sampled from visible images of the query identity $y_i$. The gallery contains RGB images $\mathcal{G}=\{V_j^{R}\}_{j=1}^{N_g}$. The objective is to rank same-identity gallery images ahead of other identities by cross-modal similarity. Gallery identity labels are unavailable to the model at inference.

\subsection{From Textual Difference to Visual Complementarity}
\label{sec3.2}

Motivated by multi-modal analogical reasoning, direct residual fusion uses $\Delta\mathbf{t}_i=\phi_T(T_i^{R})-\phi_T(T_i^{I})$, where $\phi_T(\cdot)$ denotes the text encoder, to compensate the infrared visual representation. This assumes that the difference between the descriptions corresponds to missing visual evidence.

However, dense text embeddings may jointly encode appearance attributes, linguistic patterns, and contextual information. Under semantic superposition, these factors can occupy overlapping feature directions, causing the residual to mix useful appearance changes with redundant or identity-irrelevant semantics. As illustrated in Fig.~\ref{fig1}, direct fusion may therefore introduce interference and perturb the identity-discriminative structure of the infrared representation.

Textual difference describes how two descriptions differ, whereas visual complementarity concerns whether that difference compensates for missing appearance evidence of the target identity. Neither dense subtraction nor sparsity alone guarantees this correspondence. This motivates SARC to select textual variations under semantic superposition and calibrate the resulting residual against the identity-level RGB--IR visual discrepancy observed during training.

\section{Superposition-Aware Residual Calibration}
\label{sec4}

\subsection{Overview}
\label{4.1}

As discussed in Sec.~\ref{sec3.2}, textual differences do not necessarily correspond to the visual information missing from infrared observations. Under semantic superposition, dense subtraction may mix useful appearance cues with redundant or identity-irrelevant semantics. Reliable residual fusion therefore requires both selective semantic transfer and identity-specific visual calibration.

We propose \textbf{SARC} (\textbf{S}uperposition-\textbf{A}ware \textbf{R}esidual \textbf{C}alibration) to address these requirements. Fig.~\ref{fig2} illustrates its overall architecture. SARC decomposes multimodal representations into sparse feature strengths over a shared dictionary of learnable directions. It selects salient strength variations between RGB- and IR-oriented descriptions and reconstructs a textual residual while suppressing interference among overlapping directions. The residual is then calibrated against the identity-level RGB--IR visual discrepancy observed during training. Cross-identity counterfactual residuals provide additional supervision to distinguish identity-specific compensation from generic modality shifts. Together, these components address two complementary questions: which textual variations can be transferred with limited interference, and whether they provide visual compensation consistent with the target identity.

\subsection{Low-Interference Sparse Residual Learning}
\label{4.2}

Let $\phi_R(\cdot)$ and $\phi_I(\cdot)$ denote the RGB and infrared image encoders, respectively, and let $\phi_T(\cdot)$ denote the shared text encoder. For the $i$-th sample, we obtain visual representations $\mathbf{v}_i^R=\phi_R(V_i^R)$ and $\mathbf{v}_i^I=\phi_I(V_i^I)$, and textual representations $\mathbf{t}_i^R=\phi_T(T_i^R)$ and $\mathbf{t}_i^I=\phi_T(T_i^I)$. All representations lie in a shared $d$-dimensional embedding space.

Under semantic superposition, dense textual differences may combine appearance-related variations with linguistic and contextual factors, making individual semantic contributions difficult to isolate and control. To enable more structured residual construction, we introduce a shared sparse encoder $E(\cdot)$ and a learnable dictionary $\mathbf{W}=[\mathbf{w}_1,\ldots,\mathbf{w}_K] \in\mathbb{R}^{d\times K}$. The dictionary provides a common set of feature directions for representing RGB-oriented and IR-oriented textual embeddings, allowing their differences to be expressed as changes in activation strength along corresponding directions.

For a generic input representation $\mathbf{t}\in\mathbb{R}^{d}$, the encoder produces non-negative activations, from which at most $k$ entries are retained through Top-$k$ sparsification:
\begin{equation}
    \begin{aligned}
        \mathbf{X}
        &=\operatorname{TopK}_{k}(E(\mathbf{t}))
        \in\mathbb{R}_{+}^{K},\\
        \tilde{\mathbf{t}}
        &=\mathbf{W}\mathbf{X}
        =\sum_{j=1}^{K}x_j\mathbf{w}_j,
    \end{aligned}
\end{equation}
where $\mathbf{X}=[x_1,\ldots,x_K]^{\top}$ and $x_j$ denotes the activation strength associated with the learned direction $\mathbf{w}_j$.

The encoder and dictionary are shared across $\mathbf{v}^{R}$, $\mathbf{v}^{I}$, $\mathbf{t}^{R}$, and $\mathbf{t}^{I}$, so that each coordinate refers to the same learned direction and uses the same activation mapping across these representations. This establishes the coordinate correspondence needed for feature-strength comparison: the $j$-th strengths of the two descriptions both weight $\mathbf{w}_j$. Independently learned dictionaries would not ensure this correspondence, since their coordinate ordering and learned directions could differ. In particular, RGB-oriented and IR-oriented textual embeddings are independently encoded and sparsified before their feature strengths are compared. This shared-coordinate formulation makes it possible to identify and selectively retain variations along corresponding directions, rather than treating the entire dense textual difference as an inseparable residual. Together with subsequent low-interference constraints, it provides a structured basis for suppressing overlapping residual contributions. In implementation, $E(\cdot)$ consists of a single $d$-to-$K$ affine layer followed by ReLU, while Top-$k$ selection is applied separately to its output.

We learn the shared decomposition by balancing reconstruction fidelity with reduced dictionary overlap. The reconstruction objective covers all four types of representations:
\begin{equation}
    \mathcal{L}_{\mathrm{rec}}
    =
    \mathbb{E}_{\mathbf{t}}
    \left[
        \|\mathbf{t}-\tilde{\mathbf{t}}\|_2^2
    \right].
\end{equation}
Since sparsity alone does not prevent dictionary vectors from converging to similar directions, we normalize each nonzero vector as $\bar{\mathbf{w}}_j=\mathbf{w}_j/\|\mathbf{w}_j\|_2$ and penalize excessive pairwise correlation:
\begin{equation}
    \mathcal{L}_{\mathrm{coh}}
    =
    \frac{2}{K(K-1)}
    \sum_{1\le j<l\le K}
    \left(
    \bar{\mathbf{w}}_j^{\top}\bar{\mathbf{w}}_l
    \right)^2.
\end{equation}
The resulting dictionary objective, $\mathcal{L}_{\mathrm{dict}} =\mathcal{L}_{\mathrm{rec}}+\lambda_{1}\mathcal{L}_{\mathrm{coh}}$, encourages informative sparse representations with less redundancy among feature directions

Given the two descriptions, we obtain their sparse strengths as $\mathbf{f}_i^R=\operatorname{TopK}_{k}(E(\mathbf{t}_i^R))$ and $\mathbf{f}_i^I=\operatorname{TopK}_{k}(E(\mathbf{t}_i^I))$. A positive strength difference indicates a direction emphasized more strongly by the RGB-oriented description, making it a candidate for additional visible appearance evidence. We retain these increments and modulate them with a learnable gate:
\begin{equation}
    \mathbf{s}_i
    =
    \mathbf{g}\odot
    \operatorname{ReLU}
    \left(
        \mathbf{f}_i^R-\mathbf{f}_i^I
    \right),
\end{equation}
where $\mathbf{g}$ is a global learnable vector and $\mathbf{s}_i$ is the selected strength residual. The gate is obtained by applying sigmoid to $K$ shared trainable logits initialized to zero, so each gate value initially equals $0.5$. We retain only positive variations because the residual is intended to supplement semantic evidence that is stronger in the RGB-oriented representation than in the infrared-oriented representation. The sample-dependent difference captures textual variation, while the gate learns the contribution of each feature direction across training samples. Positive increments provide a directional selection rule rather than a guarantee of visual complementarity; their relevance to identity matching is further assessed by retrieval supervision and subsequent visual calibration. Top-$k$ is applied separately within each encoder before subtraction. Since both codes are non-negative, a positive difference can occur only at an active coordinate of $\mathbf{f}_i^R$, so at most $k$ entries remain. If fewer than $k$ positive entries survive, they are retained without adding nonzero entries; an all-zero difference produces a zero residual.

Global dictionary overlap affects a residual only through the directions that it activates jointly. We therefore complement $\mathcal{L}_{\mathrm{coh}}$ with an activation-dependent interference penalty. Let $C_{jl}=|\bar{\mathbf{w}}_j^{\top}\bar{\mathbf{w}}_l|$ for $j\neq l$, and $C_{jj}=0$. For a mini-batch of size $B$, we define
\begin{equation}
    \mathcal{L}_{\mathrm{int}}
    =
    \frac{1}{B}
    \sum_{i=1}^{B}
    \frac{
        |\mathbf{s}_i|^{\top}\mathbf{C}|\mathbf{s}_i|
    }{
        \|\mathbf{s}_i\|_1^2-\|\mathbf{s}_i\|_2^2+\epsilon
    },
\end{equation}
where $|\cdot|$ denotes the elementwise absolute value and $\epsilon=10^{-8}$ is used for numerical stability. In implementation, samples whose pairwise activation mass $\|\mathbf{s}_i\|_1^2-\|\mathbf{s}_i\|_2^2$ does not exceed $\epsilon$ are skipped, and the loss is averaged over the remaining samples with the denominator floored at $\epsilon$. If none remain, the loss is zero. This excludes zero- and single-entry residuals. The numerator weights directional overlap by the magnitudes of jointly active strengths, while the denominator normalizes by their total pairwise activation mass, reducing sensitivity to residual scale. Thus, $\mathcal{L}_{\mathrm{coh}}$ regularizes dictionary geometry globally, whereas $\mathcal{L}_{\mathrm{int}}$ targets the overlap expressed by each selected residual.

Together, the shared coordinates specify which feature strengths can be compared, sparse positive selection determines the candidate increments, and the two interference constraints limit overlap globally and among jointly active components. The selected strengths are reconstructed as $\mathbf{r}_i=\mathbf{W}\mathbf{s}_i$ and fused with the infrared feature:
\begin{equation}
    \mathbf{f}_i^{\mathrm{sp}}
    =
    \mathbf{v}_i^I+\alpha\mathbf{r}_i,
\end{equation}
where $\alpha$ controls the contribution of textual compensation. To connect residual selection with identity discrimination, we optimize the fused representation using identity classification and weighted regularized triplet (WRT) loss:
\begin{equation}
    \begin{aligned}
        \mathcal{L}_{\mathrm{id}}
        &=
        -\frac{1}{B}
        \sum_{i=1}^{B}
        \log p(y_i\mid\mathbf{f}_i^{\mathrm{sp}}), \\
        \mathcal{L}_{\mathrm{wrt}}
        &=
        \frac{1}{B}
        \sum_{i=1}^{B}
        \operatorname{softplus}
        \left(d_i^{+}-d_i^{-}\right).
    \end{aligned}
\end{equation}
Here, $p(y_i\mid\mathbf{f}_i^{\mathrm{sp}})$ is the predicted probability of the ground-truth identity, and $d_i^{+}$ and $d_i^{-}$ are weighted positive and negative distances that emphasize distant positives and nearby negatives, respectively.

\subsection{Visual Complement and Counterfactual Calibration}
\label{4.3}

Although $\mathbf{r}_i$ is sparse and interference-aware, it is still derived from a textual difference and therefore cannot be assumed to be a genuine visual complement. SARC consequently uses identity-specific RGB--IR discrepancies available during training as supervisory signals. For identity $y$, we first compute modality-specific prototypes:
\begin{equation}
    \mathbf{p}_y^m
    =
    \operatorname{Norm}
    \left(
        \frac{1}{|\mathcal{I}_y^m|}
        \sum_{i\in\mathcal{I}_y^m}
        \operatorname{Norm}(\mathbf{v}_i^m)
    \right),
    m\in\{R,I\},
\end{equation}
where $\mathcal{I}_y^m$ denotes the training samples of identity $y$ in modality $m$, and $\operatorname{Norm}(\mathbf{x})=\mathbf{x}/\|\mathbf{x}\|_2$. The corresponding visual complement direction is
\begin{equation}
    \mathbf{c}_y
    =
    \operatorname{Norm}
    \left(
        \mathbf{p}_y^R-\mathbf{p}_y^I
    \right).
\end{equation}
Identity-level aggregation reduces the influence of pose, viewpoint, background, and imperfect cross-modal correspondence. Importantly, $\mathbf{c}_y$ is constructed exclusively from training identities and is used only to provide supervision during training. It is not required at inference, where the identity labels and RGB--IR prototypes of the query are unavailable. Since the training and test identity sets are disjoint under the standard ReID protocol, the learned calibration must generalize beyond the identities used to construct these prototypes.

Instead of directly replacing the textual residual with $\mathbf{c}_y$, which would discard sample-specific information, SARC learns a lightweight adapter $A(\cdot)$ to refine the residual:
\begin{equation}
    \begin{aligned}
        \tilde{\mathbf{r}}_i
        &=
        \mathbf{r}_i+\eta A(\mathbf{r}_i), \\
        \mathbf{f}_i
        &=
        \mathbf{v}_i^I+\alpha\tilde{\mathbf{r}}_i.
    \end{aligned}
\end{equation}
Distinct from the gate $\mathbf{g}$, $A(\cdot)$ is a two-layer bottleneck MLP ($d\rightarrow128\rightarrow d$), with input LayerNorm and a GELU activation between the linear layers. Its output layer is initialized to zero, so calibration starts from the sparse-fusion representation $\mathbf{f}_i^{\mathrm{sp}}$. During training, the refined residual is encouraged to agree with the visual complement of its identity:
\begin{equation}
    \mathcal{L}_{\mathrm{comp}}
    =
    \frac{1}{B}
    \sum_{i=1}^{B}
    \left[
        1-
        \operatorname{sim}
        \left(
            \tilde{\mathbf{r}}_i,
            \mathbf{c}_{y_i}
        \right)
    \right],
\end{equation}
where $\operatorname{sim}(\cdot,\cdot)$ denotes cosine similarity. Thus, the prototype direction serves as a training-time target that teaches the adapter how textual residuals should be corrected; it is not used as an additional feature during testing.

Prototype alignment alone may encourage a generic visible-modality shift. We therefore introduce a counterfactual constraint using residuals from other identities. For training query $i$, the most confusing cross-identity residual is selected as
\begin{equation}
    j_i^{-}
    =
    \arg\max_{j:\,y_j\neq y_i}
    \operatorname{sim}
    \left(
        \tilde{\mathbf{r}}_j,
        \mathbf{p}_{y_i}^{R}
    \right),
\end{equation}
and produces the identity-mismatched feature
\begin{equation}
    \mathbf{f}_i^{-}
    =
    \mathbf{v}_i^I+\alpha\tilde{\mathbf{r}}_{j_i^{-}}.
\end{equation}
The correct residual should improve the compatibility between the infrared feature and its corresponding RGB prototype more than this plausible but identity-mismatched alternative. We impose
\begin{equation}
    \mathcal{L}_{\mathrm{cf}}
    =
    \frac{1}{B}
    \sum_{i=1}^{B}
    \operatorname{softplus}
    \left(
        m
        -\operatorname{sim}
        \left(
            \mathbf{f}_i,\mathbf{p}_{y_i}^{R}
        \right)
        +\operatorname{sim}
        \left(
            \mathbf{f}_i^{-},\mathbf{p}_{y_i}^{R}
        \right)
    \right).
\end{equation}

At inference time, the learned adapter is applied directly to the sample-specific residual $\mathbf{r}_i$. No identity label, identity prototype, visual complement direction, or cross-identity residual is required. Consequently, the training-time visual discrepancy acts as a calibration teacher, while the deployed model performs residual correction using only the available query features.

\subsection{Overall Objective and Inference}
\label{4.4}

SARC follows a progressive training strategy. First, the shared sparse encoder and dictionary are learned by minimizing $\mathcal{L}_{\mathrm{dict}} =\mathcal{L}_{\mathrm{rec}}+\lambda_1\mathcal{L}_{\mathrm{coh}}$. With the encoder and dictionary fixed, the gating and sparse residual fusion components are then optimized using $\mathcal{L}_{\mathrm{sparse}} =\mathcal{L}_{\mathrm{id}} +\lambda_2\mathcal{L}_{\mathrm{wrt}} +\lambda_3\mathcal{L}_{\mathrm{int}}$. Finally, visual complement and counterfactual supervision guide residual calibration through $\mathcal{L}_{\mathrm{cal}}=\mathcal{L}_{\mathrm{id}}+\lambda_2\mathcal{L}_{\mathrm{wrt}}+\lambda_4\mathcal{L}_{\mathrm{comp}}+\lambda_5\mathcal{L}_{\mathrm{cf}}$, where the coefficients balance the respective objectives. This final stage encourages the calibrated residual to align with the visual complement of the corresponding identity while providing greater retrieval utility than residuals from other identities.

Identity prototypes and counterfactual residuals are used exclusively for training supervision. At inference, SARC encodes the RGB- and IR-oriented descriptions, constructs their gated sparse strength difference, and reconstructs the textual residual through the shared dictionary. The learned adapter then calibrates this residual before it is fused with the infrared query representation. The resulting feature is compared with RGB gallery representations for retrieval, without requiring identity labels, prototype matching, or counterfactual residual search.

\section{Experiments}
\label{sec5}

\subsection{Experimental Setup}
\label{sec5.1}

\textbf{Datasets and evaluation.}
We evaluate SARC on Tri-SYSU-MM01, Tri-RegDB, and Tri-LLCM, the text-augmented versions of SYSU-MM01~\cite{wu2017rgb}, RegDB~\cite{nguyen2017person}, and LLCM~\cite{zhang2023diverse}. Following TVI-LFM~\cite{hu2024empowering}, we evaluate both image-only queries and fusion queries containing an infrared image, its IR-oriented description, and an identity-associated RGB-oriented description. We report Rank-1, mAP, and mINP~\cite{ye2021deep}, averaged over ten gallery sampling trials for Tri-SYSU-MM01 and Tri-LLCM, and ten standard splits for Tri-RegDB.

\textbf{Implementation details.}
We use an ImageNet-pretrained dual-stream ResNet-50~\cite{ye2020dynamic,russakovsky2015imagenet} as the visual backbone and a CLIP text encoder~\cite{radford2021learning} for textual representation. The data augmentation settings are consistent with those used in TVI-LFM~\cite{hu2024empowering}. SARC is implemented in PyTorch and trained on a single NVIDIA RTX 4090 GPU. The representation dimension, dictionary size, and sparse activation budget are set to $d=2048$, $K=8192$, and $k=64$, respectively. Training proceeds in three stages: dictionary learning, sparse residual learning, and visual complement calibration. All stages use the Adam optimizer with a weight decay of $10^{-5}$. Their batch sizes are set to $(256,96,96)$, the initial learning rates to $(3\times10^{-4}, 10^{-3}, 5\times10^{-5})$, and the numbers of training epochs to $(50,30,8)$, respectively. After dictionary learning, the sparse encoder and dictionary are frozen throughout the subsequent stages. The loss weights are set to $(\lambda_1,\lambda_2,\lambda_3,\lambda_4,\lambda_5)=(0.2,0.5,0.02,0.02,0.05)$ for coherence regularization, WRT loss, interference suppression, visual complement alignment, and counterfactual supervision, respectively. We further set the residual fusion coefficient to $\alpha=0.25$, the adapter correction scale to $\eta=0.05$, and the counterfactual margin to $m=0.05$. Identity prototypes are constructed exclusively from training samples and are used only to provide training supervision.

\textbf{Complementarity metrics.}
We further introduce three metrics to evaluate whether the textual residual provides meaningful visual complementarity: Visual Complement Alignment (VCA), Discriminative Complement Gain (DCG), and Discriminative Complement Ratio (DCR). Let $\mathbf{f}_i^I$ and $\mathbf{f}_i^F$ denote the $\ell_2$-normalized infrared and fused query features, respectively, and let $\mathbf{f}_j^R$ denote the normalized feature of the $j$-th RGB gallery image. For each identity $y$, we construct an RGB prototype $\mathbf{p}_y^R$ by averaging its normalized RGB gallery features, followed by $\ell_2$ normalization. Ground-truth gallery identities are used only for this offline analysis and are never involved in feature construction or retrieval ranking.

To evaluate whether fusion moves the infrared representation toward its identity-specific RGB appearance, we define the fusion-induced displacement as $\Delta\mathbf{f}_i=\mathbf{f}_i^F-\mathbf{f}_i^I$ and the reference visual-complement direction as $\mathbf{c}_i=\mathbf{p}_{y_i}^R-\mathbf{f}_i^I$. VCA measures their directional agreement as $\mathrm{VCA}=\frac{1}{N}\sum_{i=1}^{N} \cos(\Delta\mathbf{f}_i,\mathbf{c}_i)$, where $N$ is the number of evaluated queries and $\cos(\cdot,\cdot)$ denotes cosine similarity. A higher VCA indicates that the fusion-induced change is better aligned with the identity-specific RGB--IR visual complement.

We further evaluate whether such alignment leads to improved identity discrimination. For each query, we select the hardest negative RGB gallery sample using the original infrared feature as $j_i^-=\arg\max_{j:y_j\neq y_i} \cos(\mathbf{f}_i^I,\mathbf{f}_j^R)$ and keep this sample fixed when comparing the original and fused representations. The identity margin of a query feature $\mathbf{f}$ is defined as $M_i(\mathbf{f})= \cos(\mathbf{f},\mathbf{p}_{y_i}^R)- \cos(\mathbf{f},\mathbf{f}_{j_i^-}^R)$. The margin improvement induced by fusion is then $\Delta M_i=M_i(\mathbf{f}_i^F)-M_i(\mathbf{f}_i^I)$. Based on $\Delta M_i$, DCG measures the average magnitude of discriminative improvement as $\mathrm{DCG}=\frac{1}{N}\sum_{i=1}^{N}\Delta M_i$, whereas DCR measures the proportion of queries with positive margin gains as $\mathrm{DCR}=\frac{1}{N}\sum_{i=1}^{N}\mathbb{I}[\Delta M_i>0]$, where $\mathbb{I}[\cdot]$ denotes the indicator function. VCA and DCR are reported as percentages. Together, VCA, DCG, and DCR characterize the directional consistency, average discriminative gain, and coverage of beneficial fusion, respectively.

\begin{table*}[t]
\centering
\scriptsize
\caption{Comparison with the state-of-the-art methods on Tri-SYSU-MM01. For SARC, results are reported as mean $\pm$ standard deviation over three independent runs. The best and second-best results are highlighted in \textbf{bold} and \underline{underline}, respectively.}
\label{tab1}
\setlength{\tabcolsep}{3.5pt}
\begin{tabular}{lcccccccc}
\toprule
\multirow{2}{*}{Methods} & \multirow{2}{*}{Venue} & \multirow{2}{*}{Type} & \multicolumn{3}{c}{All Search} & \multicolumn{3}{c}{Indoor Search} \\
\cmidrule(lr){4-6} \cmidrule(lr){7-9} & & & Rank-1 & mAP & mINP & Rank-1 & mAP & mINP \\
\midrule
Zero-Padding~\cite{wu2017rgb}      & ICCV-17            & \multirow{14}{*}{$I \rightarrow R$}     & 14.80 & 15.95 & --    & 20.58 & 26.92 & --    \\
HCML~\cite{ye2018hierarchical}     & AAAI-18            &                                         & 14.32 & 16.16 & --    & 24.52 & 30.08 & --    \\
cmGAN~\cite{dai2018cross}          & IJCAI-18           &                                         & 26.97 & 27.80 & --    & 31.63 & 42.19 & --    \\
AlignGAN~\cite{wang2019rgb}        & ICCV-19            &                                         & 42.40 & 40.70 & --    & 45.90 & 54.30 & --    \\
DDAG~\cite{ye2020dynamic}          & ECCV-20            &                                         & 54.75 & 53.02 & 39.62 & 61.02 & 67.98 & 62.61 \\
AGW~\cite{ye2021deep}              & TPAMI-21           &                                         & 47.50 & 47.65 & 35.30 & 54.17 & 62.97 & 59.23 \\
CM-NAS~\cite{fu2021cm}             & ICCV-21            &                                         & 61.99 & 60.02 & --    & 67.01 & 72.95 & --    \\
CAJ~\cite{ye2021channel}           & ICCV-21            &                                         & 69.88 & 66.89 & 53.61 & 76.26 & 80.37 & 76.79 \\
DART~\cite{yang2022learning}       & CVPR-22            &                                         & 68.72 & 66.29 & 53.26 & 72.52 & 78.17 & 74.94 \\
MSCLNet~\cite{zhang2022modality}   & ECCV-22            &                                         & 76.99 & 71.64 & --    & 78.49 & 81.17 & --    \\
DEEN~\cite{zhang2023diverse}       & CVPR-23            &                                         & 74.70 & 71.80 & --    & 80.30 & 83.30 & --    \\
SAAI~\cite{fang2023visible}        & ICCV-23            &                                         & 75.90 & 77.03 & --    & 83.20 & 88.01 & --    \\
SGIEL~\cite{feng2023shape}         & CVPR-23            &                                         & 77.12 & 72.33 & --    & 82.07 & 82.95 & --    \\
PartMix~\cite{kim2023partmix}      & CVPR-23            &                                         & 77.78 & 74.62 & --    & 81.52 & 84.38 & --    \\
\midrule
YYDS~\cite{du2024yyds}             & arXiv-24           & \multirow{11}{*}{$I + T \rightarrow R$} & 74.60 & 70.35 & 56.01 & 81.35 & 83.64 & 79.56 \\
TVI-LFM~\cite{hu2024empowering}    & NeurIPS-24         &                                         & \textbf{84.90} & \underline{81.47} & \underline{70.85} & \underline{89.06} & \underline{90.78} & \underline{88.39} \\
STAR-ReID~\cite{qiu2025advancing}  & TIFS-25            &                                         & 82.93 & 80.47 & --    & 88.04 & 89.58 & --    \\
DSFAD~\cite{dong2025diverse}       & TIFS-25            &                                         & 78.40 & 74.70 & --    & 85.10 & 87.20 & --    \\
RSBA~\cite{dong2025richer}         & IJCAI-25           &                                         & 78.40 & 74.80 & --    & 87.70 & 87.80 & --    \\
MPT~\cite{zhang2025visible}        & ICMR-25            &                                         & 78.64 & 74.22 & --    & 83.12 & 85.03 & --    \\
CSDN~\cite{yu2025clip}             & TMM-25             &                                         & 76.70 & 73.00 & --    & 84.50 & 86.80 & --    \\
CDSA~\cite{lu2026clip}             & IEEE MM-26         &                                         & 75.00 & 72.40 & --    & 83.90 & 86.90 & --    \\
PBSA-Net~\cite{huang2026clip}      & Algorithms-26      &                                         & 76.50 & 74.20 & --    & 84.10 & 87.20 & --    \\
VG-MPL~\cite{luo2026visible}       & ESWA-26            &                                         & 77.43 & 75.68 & --    & 86.42 & 88.22 & --    \\
CLIP-SIF~\cite{liu2026visible}     & ICSP-26            &                                         & 76.20 & 73.10 & --    & 82.30 & 85.50 & --    \\
\midrule
VI-ReID Backbone                   & --                 & $I \rightarrow R$                       & 68.23 & 65.84 & 52.62 & 76.63 & 80.74 & 77.10 \\
SARC                               & --                 & $I + T \rightarrow R$                   & $\underline{84.80 \pm 0.05}$ & $\mathbf{81.98 \pm 0.02}$ & $\mathbf{71.92 \pm 0.08}$ & $\mathbf{89.64 \pm 0.43}$ & $\mathbf{91.30 \pm 0.19}$ & $\mathbf{89.10 \pm 0.17}$ \\
\bottomrule
\end{tabular}
\end{table*}

\begin{table*}[t]
\centering
\scriptsize
\caption{Comparison with the state-of-the-art methods on Tri-RegDB and Tri-LLCM. For SARC, results are reported as mean $\pm$ standard deviation over three independent runs. The best and second-best results are highlighted in \textbf{bold} and \underline{underline}, respectively.}
\label{tab2}
\setlength{\tabcolsep}{3.5pt}
\begin{tabular}{lcccccccc}
\toprule
\multirow{2}{*}{Methods} & \multirow{2}{*}{Venue} & \multirow{2}{*}{Type} & \multicolumn{3}{c}{Tri-RegDB} & \multicolumn{3}{c}{Tri-LLCM} \\
\cmidrule(lr){4-6} \cmidrule(lr){7-9} & & & Rank-1 & mAP & mINP & Rank-1 & mAP & mINP \\
\midrule
DDAG~\cite{ye2020dynamic}          & ECCV-20            & \multirow{6}{*}{$I \rightarrow R$}     & 68.06 & 61.80 & 48.62 & 40.30 & 48.40 & --    \\
AGW~\cite{ye2021deep}              & TPAMI-21           &                                        & 70.49 & 65.90 & 51.24 & 43.60 & 51.80 & --    \\
CAJ~\cite{ye2021channel}           & ICCV-21            &                                        & 84.80 & 77.80 & 61.56 & 48.80 & 56.60 & --    \\
MMN~\cite{zhang2021towards}        & ACM MM-21          &                                        & 87.50 & 80.50 & --    & 52.50 & 58.90 & --    \\
DART~\cite{yang2022learning}       & CVPR-22            &                                        & 81.97 & 73.78 & 56.70 & --    & --    & --    \\
DEEN~\cite{zhang2023diverse}       & CVPR-23            &                                        & 89.50 & 83.40 & --    & 54.90 & 62.90 & --    \\
\midrule
YYDS~\cite{du2024yyds}             & arXiv-24           & \multirow{8}{*}{$I + T \rightarrow R$} & 90.95 & 84.22 & 70.12 & 58.13 & 64.91 & 61.77 \\
TVI-LFM~\cite{hu2024empowering} & NeurIPS-24         &                                        & 91.38 & 85.92 & 72.73 & \underline{58.19} & \underline{65.08} & \underline{61.83} \\
STAR-ReID~\cite{qiu2025advancing}  & TIFS-25            &                                        & 90.48 & \textbf{91.77} & --    & --    & --    & --    \\
CSDN~\cite{yu2025clip}             & TMM-25             &                                        & \textbf{92.30} & 85.50 & --    & 55.80 & 63.50 & --    \\
CDSA~\cite{lu2026clip}             & IEEE MM-26         &                                        & 78.70 & 73.30 & --    & 52.50 & 63.30 & --    \\
PBSA-Net~\cite{huang2026clip}      & Algorithms-26      &                                        & 78.90 & 73.90 & --    & 53.70 & 64.20 & --    \\
MCST~\cite{xiong2026clip}          & PR-26              &                                        & 87.80 & 84.40 & \textbf{74.50} & 55.20 & 62.50 & 58.30  \\
CLIP-SIF~\cite{liu2026visible}     & ICSP-26            &                                        & 88.10 & 83.50 & --    & --    & --    & --    \\
\midrule
VI-ReID Backbone                   & --                 & $I \rightarrow R$                      & 87.86 & 82.20 & 67.89 & 52.96 & 59.41 & 56.06 \\
SARC                               & --                 & $I + T \rightarrow R$                  & $\underline{91.70 \pm 0.73}$ & $\underline{86.73 \pm 0.74}$ & $\underline{74.41 \pm 0.77}$ & $\mathbf{59.57 \pm 0.13}$ & $\mathbf{66.10 \pm 0.10}$ & $\mathbf{62.99 \pm 0.10}$ \\
\bottomrule
\end{tabular}
\end{table*}

\subsection{Comparison with State-of-the-Art Methods}
\label{sec5.2}

We compare SARC with representative state-of-the-art VI-ReID methods on Tri-SYSU-MM01, Tri-RegDB, and Tri-LLCM. The compared approaches are grouped into image-only methods ($I \rightarrow R$) and image-text-assisted methods ($I+T \rightarrow R$), according to whether textual information is involved in representation learning. The results of the compared methods are taken from their original publications. For SARC, we report the mean and standard deviation over three independent runs. The results are summarized in Tables~\ref{tab1} and~\ref{tab2}.

As shown in Table~\ref{tab1}, SARC achieves the best performance in five of the six evaluation metrics under the All Search and Indoor Search protocols on Tri-SYSU-MM01. Under All Search, SARC obtains 84.80\% Rank-1, 81.98\% mAP, and 71.92\% mINP, outperforming TVI-LFM by 0.51 and 1.07 percentage points (pp) in mAP and mINP, respectively, while achieving comparable Rank-1 accuracy (84.80\% vs. 84.90\%). Under Indoor Search, SARC achieves the best results across all three metrics, reaching 89.64\% Rank-1, 91.30\% mAP, and 89.10\% mINP, surpassing TVI-LFM by 0.58, 0.52, and 0.71 pp, respectively. The standard deviations remain below 0.5 pp across all six metrics, indicating relatively stable performance over three independent runs. Compared with the image-only VI-ReID backbone, SARC further improves Rank-1, mAP, and mINP by 16.57, 16.14, and 19.30 pp under All Search, respectively.

Table~\ref{tab2} further presents the comparisons on Tri-RegDB and Tri-LLCM. On Tri-RegDB, SARC achieves 91.70\% Rank-1, 86.73\% mAP, and 74.41\% mINP, ranking second across all three metrics. The best Rank-1, mAP, and mINP results are obtained by CSDN, STAR-ReID, and MCST, respectively. SARC trails the best-performing methods by only 0.60 pp in Rank-1 and 0.09 pp in mINP, although a larger gap of 5.04 pp remains in mAP compared with STAR-ReID. Nevertheless, SARC consistently improves upon the image-only backbone by 3.84, 4.53, and 6.52 pp in Rank-1, mAP, and mINP, respectively. The larger gain in mINP suggests that the learned complementary information is particularly beneficial for retrieving challenging cross-modal matches. The standard deviations of 0.73, 0.74, and 0.77 pp also indicate greater run-to-run variability on Tri-RegDB than on the other two benchmarks.

On Tri-LLCM, SARC achieves the best performance across all three metrics, reaching 59.57\% Rank-1, 66.10\% mAP, and 62.99\% mINP. Compared with the second-best TVI-LFM, SARC improves Rank-1, mAP, and mINP by 1.38, 1.02, and 1.16 pp, respectively. Relative to the image-only backbone, the corresponding gains are 6.61, 6.69, and 6.93 pp. Moreover, the standard deviations remain at or below 0.13 pp across all three metrics, indicating consistent performance over independent runs. These improvements demonstrate the effectiveness of SARC in enhancing both top-ranked retrieval accuracy and overall cross-modal matching quality.

Overall, SARC achieves the best results in five of the six metrics on Tri-SYSU-MM01 and all three metrics on Tri-LLCM, while ranking second across all three metrics on Tri-RegDB. It also consistently outperforms its image-only backbone across all benchmarks. These results support the effectiveness of extracting and calibrating complementary information from textual residuals to enhance identity-discriminative cross-modal representations.

\begin{table*}[t]
\centering
\scriptsize
\caption{Analysis of textual residual complementarity. Textual residuals are progressively refined using Low-Interference Sparse Residual Learning (LISR) and Visual Complement and Counterfactual Calibration (VCCC). VCA measures visual complement alignment, while DCG and DCR measure the magnitude and coverage of discriminative complement gains, respectively. VCA, DCR, and retrieval metrics are reported as percentages; DCG is a dimensionless margin gain. Higher values indicate better performance.}
\label{tab3}
\setlength{\tabcolsep}{3.5pt}

\begin{tabular}{lcccccccccccccc}
\toprule

\multirow{2}{*}{Method}
& \multicolumn{2}{c}{Components}
& \multicolumn{6}{c}{Tri-SYSU-MM01}
& \multicolumn{6}{c}{Tri-LLCM} \\

\cmidrule(lr){2-3}
\cmidrule(lr){4-9}
\cmidrule(lr){10-15}
& LISR & VCCC & VCA & DCG & DCR & Rank-1 & mAP & mINP & VCA & DCG & DCR & Rank-1 & mAP & mINP \\
\midrule
IR only & -- & -- & -- & -- & -- & 68.2 & 65.8 & 52.6 & -- & -- & -- & 53.0 & 59.4 & 56.1 \\
Dense textual residual & \xmark & \xmark & 34.3 & 0.0311 & 66.6 & 84.3 & 81.4 & 71.3 & 26.5 & 0.0099 & 54.4 & 58.2 & 65.1 & 61.8 \\
Sparse textual residual & \cmark & \xmark & 36.1 & 0.0378 & 73.8 & 84.6 & 81.8 & 71.6 & 28.0 & 0.0190 & 63.3 & 59.4 & 65.9 & 62.7 \\
Calibrated textual residual & \cmark & \cmark & \textbf{36.4} & \textbf{0.0393} & \textbf{74.0} & \textbf{84.8} & \textbf{82.0} & \textbf{71.9} & \textbf{28.3} & \textbf{0.0191} & \textbf{63.9} & \textbf{59.6} & \textbf{66.1} & \textbf{63.0} \\
\bottomrule
\end{tabular}
\end{table*}

\subsection{Evaluating the Visual Complementarity of Textual Residuals}
\label{sec5.3}

Section~\ref{sec3.2} raises a central question of this work: whether the difference between modality-oriented textual representations can reliably serve as a visual complement for infrared features. To examine this question beyond conventional retrieval accuracy, we evaluate textual residuals from three complementary perspectives. VCA measures whether the fusion-induced feature displacement is directionally aligned with the prototype-based RGB--IR visual-complement direction; DCG measures the average improvement in the identity-discriminative margin; and DCR measures the proportion of queries that obtain a positive margin gain. Together, these metrics characterize \emph{where} the residual moves the representation, \emph{how much} this movement improves identity discrimination, and \emph{how consistently} the improvement occurs across queries. Table~\ref{tab3} reports these measures together with the corresponding retrieval performance.

The dense-difference baseline is an internal control variant of SARC, rather than a reproduction of the complete TVI-LFM framework. The dense textual residual already exhibits a certain degree of visual complementarity, but the results show that such complementarity cannot be taken for granted. On SYSU-MM01, direct dense residual fusion achieves a VCA of 34.3\%, a DCG of 0.0311, and a DCR of 66.6\%. On LLCM, these values decrease to 26.5\%, 0.0099, and 54.4\%, respectively. In particular, the DCR of 54.4\% indicates that the dense residual improves the discriminative margin for only slightly more than half of the evaluated queries. Nevertheless, dense residual fusion still substantially improves retrieval performance over the IR-only representation. This contrast suggests that textual differences do contain useful complementary information, while the entire dense residual does not necessarily constitute a faithful visual complement. Instead, useful appearance variations can remain entangled with redundant or visually unsupported semantics.

LISR addresses this issue by representing textual differences through sparse, selectable feature-strength variations while reducing interference among the learned directions. After applying LISR, all three complementarity measures improve consistently. On SYSU-MM01, VCA increases from 34.3\% to 36.1\%, DCG from 0.0311 to 0.0378, and DCR from 66.6\% to 73.8\%. The improvement is more pronounced on LLCM, where VCA increases from 26.5\% to 28.0\%, DCG nearly doubles from 0.0099 to 0.0190, and DCR rises by 8.9 percentage points from 54.4\% to 63.3\%. The simultaneous improvements in directional alignment, discriminative gain, and gain coverage indicate that sparsification and low-interference learning make the textual residual behave more consistently as a cross-modal complement, rather than merely reducing its dimensional support.

VCCC further addresses the remaining mismatch between sparse textual variation and the visual complement required by the infrared representation. Although LISR produces a more selective residual, the residual is still derived from the textual representation space and is therefore not guaranteed to align with the identity-specific RGB--IR visual discrepancy. VCCC explicitly calibrates this residual using identity-level visual-complement supervision and counterfactual constraints. As a result, VCA, DCG, and DCR further increase to 36.4\%, 0.0393, and 74.0\% on SYSU-MM01, and to 28.3\%, 0.0191, and 63.9\% on LLCM. The corresponding Rank-1, mAP, and mINP also reach their highest values on both datasets. Although the gains over LISR are smaller than those obtained when moving from dense to sparse residuals, their consistent direction across both complementarity and retrieval metrics supports the intended role of VCCC as a conservative calibration mechanism: it further adjusts the selected residual toward an identity-relevant visual complement through small corrections to the reconstructed residual.

These results provide a more precise answer to the question posed in Section~\ref{sec3.2}. Textual residuals can contain useful visual-complementary information, but visual complementarity is not an intrinsic property of dense textual subtraction. The progression from dense residuals to LISR and then VCCC consistently improves visual alignment, discriminative gain, gain coverage, and final retrieval performance. This evidence supports the central design principle of SARC: textual differences should first be selectively represented under semantic superposition and then calibrated against the identity-specific RGB--IR visual discrepancy, rather than being directly treated as visual complements.

\begin{table*}[t]
\centering
\scriptsize
\caption{Ablation study on residual reformulation. Feature-strength decomposition, Top-$k$ sparsification, and learned feature selection are progressively introduced. Rank-1 (\%), mAP (\%), and mINP (\%) are reported.}
\label{tab4}
\setlength{\tabcolsep}{4pt}
\begin{tabular}{lccccccccc}
\toprule
\multirow{2}{*}{Method}
& \multicolumn{3}{c}{Components}
& \multicolumn{3}{c}{Tri-SYSU-MM01}
& \multicolumn{3}{c}{Tri-LLCM} \\
\cmidrule(lr){2-4}
\cmidrule(lr){5-7}
\cmidrule(lr){8-10}

& FS & Top-$k$ & Sel.
& Rank-1 & mAP & mINP
& Rank-1 & mAP & mINP \\
\midrule

Dense difference
& \xmark & \xmark & \xmark
& 84.25 & 81.40 & 71.28
& 58.19 & 65.08 & 61.83 \\

Feature-strength difference
& \cmark & \xmark & \xmark
& 84.39 & 81.56 & 71.48
& 58.62 & 65.13 & 62.00 \\

Sparse feature-strength difference
& \cmark & \cmark & \xmark
& 84.46 & 81.63 & 71.51
& 59.12 & 65.49 & 62.27 \\

Selected sparse difference
& \cmark & \cmark & \cmark
& \textbf{84.54} & \textbf{81.67} & \textbf{71.54}
& \textbf{59.27} & \textbf{65.64} & \textbf{62.31} \\

\bottomrule
\end{tabular}
\end{table*}

\subsection{Ablation Studies}
\label{sec5.4}

\subsubsection{Ablation Study on Residual Construction}

We investigate how textual differences should be constructed before being used as residual compensation. As shown in Table~\ref{tab4}, we progressively replace direct dense subtraction with feature-strength decomposition (FS), Top-$k$ sparsification, and learned feature selection (Sel.). All variants share the same backbone, dictionary, fusion architecture, and training protocol; only the residual construction strategy is changed.

Replacing dense difference with feature-strength difference consistently improves retrieval on both benchmarks. On Tri-SYSU-MM01, it increases mAP from $81.40\%$ to $81.56\%$ and mINP from $71.28\%$ to $71.48\%$. Similar gains are observed on Tri-LLCM, indicating that decomposed feature-strength variations provide a more suitable basis for residual construction than an undifferentiated dense textual displacement.

Top-$k$ sparsification further improves performance, especially on Tri-LLCM, where mAP and mINP increase from $65.13\%$/$62.00\%$ to $65.49\%$/$62.27\%$. This suggests that restricting residual construction to a compact set of salient feature-strength variations suppresses weak or redundant textual information. Finally, learned selection achieves the best results on both benchmarks, reaching $84.54\%$ Rank-1, $81.67\%$ mAP, and $71.54\%$ mINP on Tri-SYSU-MM01, as well as $59.27\%$, $65.64\%$, and $62.31\%$ on Tri-LLCM. These progressive improvements demonstrate that textual differences are more reliably exploited after decomposition into selectable sparse feature-strength variations.

\begin{table}[t]
\centering
\scriptsize
\caption{Ablation study on low-interference learning. We evaluate the contributions of dictionary coherence regularization $\mathcal{L}_{\mathrm{coh}}$ and activation-dependent interference regularization $\mathcal{L}_{\mathrm{int}}$. Rank-1 (\%), mAP (\%), and mINP (\%) are reported.}
\label{tab5}
\setlength{\tabcolsep}{6pt}
\begin{tabular}{cccccccc}
\toprule
\multicolumn{2}{c}{Components}
& \multicolumn{3}{c}{Tri-SYSU-MM01}
& \multicolumn{3}{c}{Tri-LLCM} \\
\cmidrule(lr){1-2}
\cmidrule(lr){3-5}
\cmidrule(lr){6-8}

$\mathcal{L}_{\mathrm{coh}}$
& $\mathcal{L}_{\mathrm{int}}$
& Rank-1 & mAP & mINP
& Rank-1 & mAP & mINP \\
\midrule

\xmark & \xmark
& 84.54 & 81.67 & 71.54
& 59.27 & 65.64 & 62.31 \\

\cmark & \xmark
& 84.58 & 81.67 & 71.52
& 59.38 & 65.74 & 62.61 \\

\xmark & \cmark
& \textbf{84.64} & 81.75 & 71.54
& \textbf{59.51} & 65.83 & \textbf{62.69} \\

\cmark & \cmark
& \textbf{84.64} & \textbf{81.82} & \textbf{71.62}
& 59.43 & \textbf{65.94} & \textbf{62.69} \\

\bottomrule
\end{tabular}
\end{table}

\subsubsection{Ablation Study on Low-Interference Learning}

Table~\ref{tab5} evaluates the complementary roles of dictionary-level coherence regularization $\mathcal{L}_{\mathrm{coh}}$ and activation-dependent interference regularization $\mathcal{L}_{\mathrm{int}}$. The former constrains overlap across shared dictionary directions, while the latter accounts for their joint activation within individual residuals.

The two regularizers exhibit different effects. Using $\mathcal{L}_{\mathrm{coh}}$ alone improves all three metrics on Tri-LLCM, with smaller changes on Tri-SYSU-MM01. Using $\mathcal{L}_{\mathrm{int}}$ alone improves Rank-1 and mAP on both datasets and yields the highest Rank-1 on Tri-LLCM. These results support considering activation patterns alongside dictionary geometry: the contribution of directional overlap to residual fusion depends on which directions are jointly active.

Their combination achieves the highest mAP on both datasets, improving over the unregularized variant by 0.15 and 0.30 percentage points on Tri-SYSU-MM01 and Tri-LLCM, respectively. It also achieves the highest mINP on Tri-SYSU-MM01 and matches the highest mINP on Tri-LLCM. Although the combined setting does not further improve Tri-LLCM Rank-1 over $\mathcal{L}_{\mathrm{int}}$ alone, its higher mAP indicates better overall ranking quality. The results support a complementary design in which dictionary coherence regularization constrains the shared representation geometry, while activation-dependent regularization targets overlap expressed in each residual.

\begin{table*}[t]
\centering
\scriptsize
\caption{Ablation study on residual directionality. All variants use feature-strength decomposition, Top-$k$ sparsification, learned feature selection, and both low-interference regularizers, without visual complement calibration. Rank-1 (\%), mAP (\%), and mINP (\%) are reported.}
\label{tab12}
\setlength{\tabcolsep}{7pt}
\begin{tabular}{lcccccc}
\toprule
Residual Operator
& \multicolumn{3}{c}{Tri-SYSU-MM01}
& \multicolumn{3}{c}{Tri-LLCM} \\
\cmidrule(lr){2-4}
\cmidrule(lr){5-7}
& Rank-1 & mAP & mINP
& Rank-1 & mAP & mINP \\
\midrule
Absolute Difference 
& 83.69 & 80.75 & 70.72
& 58.91 & 65.39 & 62.25 \\

Signed Difference
& 84.61 & 81.73 & 71.44
& 59.34 & 65.81 & 62.55 \\

Positive Difference
& \textbf{84.64} & \textbf{81.82} & \textbf{71.62}
& \textbf{59.43} & \textbf{65.94} & \textbf{62.69} \\

\bottomrule
\end{tabular}
\end{table*}

\subsubsection{Effect of Residual Directionality}

We further examine the effect of residual directionality by comparing three difference operators: absolute difference $|\mathbf{f}_i^R-\mathbf{f}_i^I|$, signed difference $\mathbf{f}_i^R-\mathbf{f}_i^I$, and the proposed positive difference $\operatorname{ReLU}(\mathbf{f}_i^R-\mathbf{f}_i^I)$. To isolate the effect of the difference operator, all variants use the same feature-strength decomposition with Top-$k$ sparsification applied before differencing, learned feature selection, and both low-interference regularizers, without visual complement calibration. The positive-difference variant corresponds to the combined-regularization setting in Table~\ref{tab5}.

As shown in Table~\ref{tab12}, the absolute difference consistently performs the worst on both benchmarks. On Tri-SYSU-MM01, it achieves 83.69\% Rank-1, 80.75\% mAP, and 70.72\% mINP, which are 0.95, 1.07, and 0.90 pp lower than those of the positive difference, respectively. A similar trend is observed on Tri-LLCM, where the corresponding gaps are 0.52, 0.55, and 0.44 pp. This result suggests that discarding the sign of a feature-strength change is undesirable, since variations that are stronger in the infrared-oriented representation are also converted into positive compensation and may therefore introduce unnecessary semantic content.

The signed difference substantially improves over the absolute difference and performs close to the proposed formulation. Nevertheless, the positive difference achieves the best result for every metric on both datasets. Compared with the signed difference, it improves Rank-1/mAP/mINP by 0.03/0.09/0.18 pp on Tri-SYSU-MM01 and by 0.09/0.13/0.14 pp on Tri-LLCM. These modest gains favor the asymmetric formulation in the reported comparisons, although their statistical significance has not been established. In our setting, the residual is intended to supplement semantic evidence that is stronger in the RGB-oriented representation but weak in the infrared-oriented representation, rather than to perform unrestricted bidirectional semantic editing. Retaining only positive feature-strength variations therefore provides a simple and effective directional prior while avoiding potentially conflicting negative corrections. Overall, these results support positive differencing as a directional prior for residual construction, while signed differencing remains a competitive alternative in this setting.

\begin{table*}[t]
\centering
\scriptsize
\caption{Ablation study on visual complement target design. All gap-based variants use $\mathcal{L}_{\mathrm{comp}}$ only, while $\mathcal{L}_{\mathrm{cf}}$ is disabled to isolate the effect of target construction. Rank-1 (\%), mAP (\%), and mINP (\%) are reported.}
\label{tab6}
\setlength{\tabcolsep}{7pt}
\begin{tabular}{lcccccc}
\toprule
\multirow{2}{*}{Visual Complement Target}
& \multicolumn{3}{c}{Tri-SYSU-MM01}
& \multicolumn{3}{c}{Tri-LLCM} \\
\cmidrule(lr){2-4}
\cmidrule(lr){5-7}

& Rank-1 & mAP & mINP
& Rank-1 & mAP & mINP \\
\midrule

No calibration
& 84.64 & 81.82 & 71.62
& 59.43 & 65.94 & 62.69 \\

Identity-shuffled gap
& 84.65 & 81.82 & 71.61
& 59.45 & 65.95 & 62.69 \\

Global modality gap
& 84.66 & 81.83 & 71.61
& 59.46 & 65.96 & 62.71 \\

Instance-level gap
& 84.66 & 81.83 & 71.62
& 59.50 & \textbf{66.00} & 62.76 \\

Identity-level gap
& \textbf{84.78} & \textbf{81.97} & \textbf{71.84}
& \textbf{59.56} & \textbf{66.00} & \textbf{62.86} \\

\bottomrule
\end{tabular}
\end{table*}

\subsubsection{Ablation Study on Visual Complement Target Design}

Table~\ref{tab6} examines what constitutes a valid visual complement target for calibrating textual residuals. To isolate the effect of target construction from that of counterfactual calibration, $\mathcal{L}_{\mathrm{cf}}$ is disabled throughout this study. The gap-based variants therefore differ only in the target used by $\mathcal{L}_{\mathrm{comp}}$, while the no-calibration baseline uses neither a visual complement target nor $\mathcal{L}_{\mathrm{comp}}$. We compare an identity-shuffled gap, a global modality-level gap shared by all identities, an instance-level RGB--IR gap, and the proposed identity-level RGB--IR gap.

The identity-shuffled gap performs nearly identically to the uncalibrated baseline on both datasets. This result indicates that visual-gap supervision is ineffective when the gap is associated with an incorrect identity. Similarly, the global modality-level gap provides only marginal improvements, suggesting that a generic RGB--IR translation direction cannot adequately characterize the visual information missing from a particular infrared query.

The instance-level gap achieves slightly better results than the global modality-level target, especially on Tri-LLCM, where it improves mAP from $65.96\%$ to $66.00\%$. However, individual RGB--IR differences are affected by pose, viewpoint, background, and imperfect cross-modal correspondence. In contrast, the identity-level target achieves the best overall performance on both benchmarks, improving mAP/mINP from $81.82\%$/$71.62\%$ to $81.97\%$/$71.84\%$ on Tri-SYSU-MM01, and from $65.94\%$/$62.69\%$ to $66.00\%$/$62.86\%$ on Tri-LLCM.

These results show that effective visual complement supervision requires not only a modality-gap direction but also the correct identity-specific visual context. By aggregating RGB and IR representations at the identity level, the proposed target suppresses instance-specific nuisance variations while preserving the identity-relevant cross-modal discrepancy.

\begin{table}[t]
\centering
\scriptsize
\caption{Ablation study of the VCCC loss components. Rank-1 (\%), mAP (\%), and mINP (\%) are reported.}
\label{tab7}
\setlength{\tabcolsep}{6pt}
\begin{tabular}{cccccccc}
\toprule
\multicolumn{2}{c}{Components}
& \multicolumn{3}{c}{Tri-SYSU-MM01}
& \multicolumn{3}{c}{Tri-LLCM} \\
\cmidrule(lr){1-2}
\cmidrule(lr){3-5}
\cmidrule(lr){6-8}

$\mathcal{L}_{\mathrm{comp}}$
& $\mathcal{L}_{\mathrm{cf}}$
& Rank-1 & mAP & mINP
& Rank-1 & mAP & mINP \\
\midrule

\xmark & \xmark
& 84.64 & 81.82 & 71.62
& 59.43 & 65.94 & 62.69 \\

\cmark & \xmark
& 84.78 & 81.97 & 71.84
& 59.56 & 66.00 & 62.86 \\

\xmark & \cmark
& 84.69 & 81.85 & 71.67
& \textbf{59.64} & 66.08 & 62.94 \\

\cmark & \cmark
& \textbf{84.80} & \textbf{81.98} & \textbf{71.92}
& 59.57 & \textbf{66.10} & \textbf{62.99} \\

\bottomrule
\end{tabular}
\end{table}

\subsubsection{Ablation Study on VCCC Loss Components}

We further investigate the contributions of the two objectives in VCCC, namely the visual complement loss $\mathcal{L}_{\mathrm{comp}}$ and the counterfactual calibration loss $\mathcal{L}_{\mathrm{cf}}$. As shown in Table~\ref{tab7}, we evaluate four configurations by independently enabling the two losses while keeping all other settings unchanged.

Using $\mathcal{L}_{\mathrm{comp}}$ alone consistently improves performance on both datasets, increasing Rank-1/mAP/mINP from $84.64\%$/$81.82\%$/$71.62\%$ to $84.78\%$/$81.97\%$/$71.84\%$ on Tri-SYSU-MM01. $\mathcal{L}_{\mathrm{cf}}$ also provides consistent gains, particularly on Tri-LLCM, where Rank-1/mAP/mINP improve from $59.43\%$/$65.94\%$/$62.69\%$ to $59.64\%$/$66.08\%$/$62.94\%$. These results indicate that visual-complement alignment and counterfactual calibration provide complementary supervision for refining textual residuals.

Combining both objectives yields the best overall performance. The full VCCC achieves $84.80\%$/$81.98\%$/$71.92\%$ on Tri-SYSU-MM01 and $59.57\%$/$66.10\%$/$62.99\%$ on Tri-LLCM. Although the Rank-1 score on Tri-LLCM is slightly lower than using $\mathcal{L}_{\mathrm{cf}}$ alone, the highest mAP and mINP demonstrate improved overall ranking quality. Overall, $\mathcal{L}_{\mathrm{comp}}$ aligns textual residuals with identity-specific visual discrepancies, while $\mathcal{L}_{\mathrm{cf}}$ further constrains their contribution through counterfactual calibration, jointly producing more reliable cross-modal complementary representations.

\begin{table}[t]
\centering
\scriptsize
\caption{Top-$m$ change concentration $C_m$ (\%) under controlled textual edits on Tri-SYSU-MM01. Higher values indicate that textual variations are concentrated in fewer dimensions.}
\label{tab8}
\setlength{\tabcolsep}{7pt}
\begin{tabular}{lccccc}
\toprule
Representation
& $C_1$
& $C_4$
& $C_{16}$
& $C_{64}$
& $C_{256}$ \\
\midrule

Dense
& 0.61
& 1.94
& 5.08
& 13.45
& 35.74 \\

FS
& 0.82
& 2.60
& 7.29
& 19.31
& 49.55 \\

Sparse FS
& \textbf{3.20}
& \textbf{10.70}
& \textbf{34.37}
& \textbf{94.48}
& \textbf{100.00} \\

\bottomrule
\end{tabular}
\end{table}

\begin{figure*}[t]
    \centering
    \begin{subfigure}[b]{0.44\textwidth}
        \centering
        \includegraphics[width=\textwidth]{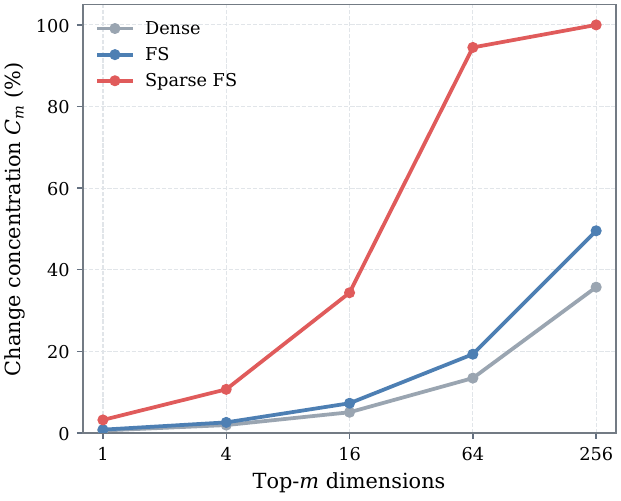}
        \caption{Top-$m$ change concentration}
        \label{fig3.1}
    \end{subfigure}
    \begin{subfigure}[b]{0.54\textwidth}
        \centering
        \includegraphics[width=\textwidth]{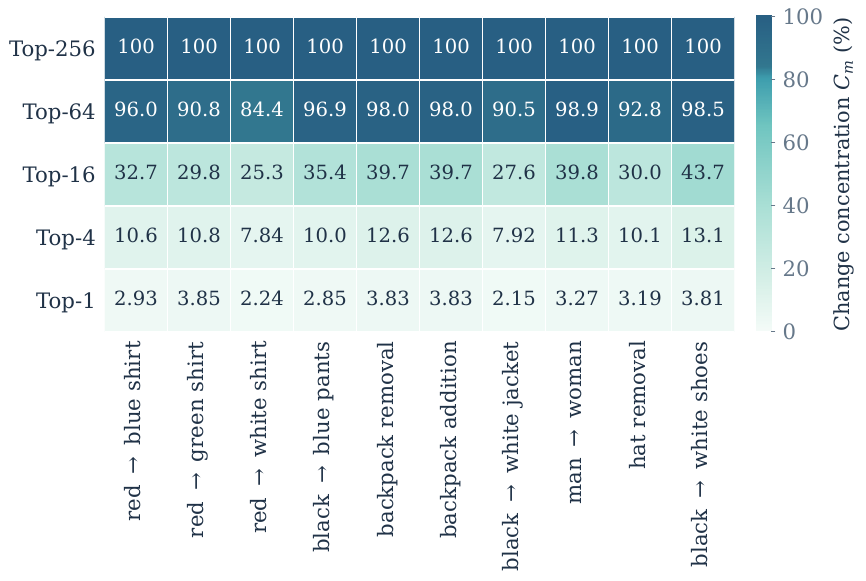}
        \caption{Per-edit concentration of Sparse FS residuals}
        \label{fig3.2}
    \end{subfigure}
    \caption{Feature-strength selectivity under controlled textual edits on Tri-SYSU-MM01. (a) Top-$m$ change concentration for dense, non-sparse feature-strength (FS), and sparse feature-strength (Sparse FS) residuals. (b) Per-edit concentration of Sparse FS residuals. Top-$k$ sparsification consistently concentrates textual changes into a compact subset of feature-strength directions.
    }
    \label{fig3}
\end{figure*}

\subsection{Mechanism Analysis}
\label{sec5.5}

\subsubsection{Feature-Strength Selectivity}

We examine whether textual changes can be expressed as compact, selectable variations in feature strength. We construct ten controlled text-edit pairs covering clothing colors, accessory presence, and gender-related wording, with each pair modifying a single attribute. For a pair $(T_a,T_b)$, let $\Delta\mathbf{h}$ denote the difference between its representations. We measure change concentration as $C_m(\Delta\mathbf{h})= \sum_{j\in\mathcal{S}_m}|\Delta h_j| / \|\Delta\mathbf{h}\|_1$, where $\mathcal{S}_m$ contains the indices of the $m$ largest absolute changes. Higher $C_m$ indicates that fewer coordinates account for a larger share of the total change magnitude. All concentration values are reported as percentages, and Table~\ref{tab8} presents averages over the ten pairs.

We compare dense textual differences (Dense), feature-strength differences before sparsification (FS), and differences between Top-$k$ sparse codes (Sparse FS). For this analysis, we use $d=K=2048$ and $k=64$ to compare the three representations at equal dimensionality. As shown in Table~\ref{tab8} and Fig.~\ref{fig3.1}, FS increases $C_{64}$ from 13.45\% to 19.31\% even before sparsification, suggesting that the learned representation redistributes textual variation toward a smaller set of coordinates. Sparse FS further concentrates 34.37\% of the change magnitude within 16 coordinates and 94.48\% within 64 coordinates. This compact representation makes individual contributions accessible to subsequent gating and residual reconstruction.

Figure~\ref{fig3.2} shows that this concentration extends across the examined edits, with Sparse FS $C_{64}$ ranging from 84.4\% to 98.9\%. The variation across edits indicates that the same activation budget can accommodate different distributions of change: some edits are dominated by fewer strength variations, whereas others spread their change more broadly. Thus, the sparse representation provides a compact set of candidate contributions without requiring every edit to activate an identical pattern.

The saturation at $C_{256}\approx100\%$ follows from the coding constraint: the difference between two codes with at most $k=64$ active entries has at most $2k=128$ nonzero entries. The more informative observations are therefore the concentration at smaller budgets and its consistency across edits. These results characterize the selectivity enabled by the learned strength representation and Top-$k$ coding; they support selective residual construction without requiring each dictionary direction to correspond to a single semantic concept.

\begin{table}[t]
\centering
\scriptsize
\caption{Dictionary geometry and active-interference analysis of LISR on Tri-SYSU-MM01. Lower values indicate reduced directional overlap and interference.}
\label{tab9}
\setlength{\tabcolsep}{8pt}
\begin{tabular}{lccc}
\toprule
Metric $\downarrow$
& w/o Low-Interference
& Full LISR
& Relative Change \\
\midrule
Coh
& 0.01837
& \textbf{0.01485}
& $-19.2\%$ \\

AI
& 0.04744
& \textbf{0.04701}
& $-0.9\%$ \\
\bottomrule
\end{tabular}
\end{table}

\subsubsection{Dictionary Geometry and Active Interference}

We examine dictionary geometry and residual-level interference using mean dictionary coherence (Coh) and active interference (AI). Following the notation introduced above, we define $\mathrm{Coh}=\frac{2}{K(K-1)} \sum_{1\leq j<l\leq K} |\bar{\mathbf{w}}_j^\top\bar{\mathbf{w}}_l|$, where $\bar{\mathbf{w}}_j$ is the normalized $j$-th dictionary direction. Coh measures average absolute directional overlap, whereas $\mathcal{L}_{\mathrm{coh}}$ penalizes squared overlap during training. We further define $\mathrm{AI}=\frac{1}{N}\sum_{i=1}^{N} \frac{|\mathbf{s}_i|^\top\mathbf{C}|\mathbf{s}_i|} {\|\mathbf{s}_i\|_1^2-\|\mathbf{s}_i\|_2^2}$, where $N$ denotes the number of valid training residual samples satisfying $\|\mathbf{s}_i\|_1^2-\|\mathbf{s}_i\|_2^2>\epsilon$, with $\epsilon=10^{-8}$, and the sum runs only over these samples, $\mathbf{s}_i$ is the selected strength residual of the $i$-th sample, $C_{jl}=|\bar{\mathbf{w}}_j^\top\bar{\mathbf{w}}_l|$ for $j\neq l$, and $C_{jj}=0$. AI averages activation-weighted interference over these samples using the same formulation as $\mathcal{L}_{\mathrm{int}}$.

Table~\ref{tab9} shows that Full LISR reduces Coh from 0.01837 to 0.01485, a relative decrease of 19.2\%. AI also decreases from 0.04744 to 0.04701, corresponding to a 0.9\% reduction. These changes indicate reduced overlap in both the shared dictionary and the directions jointly activated by residuals. Their different magnitudes reflect distinct weighting schemes: Coh treats dictionary pairs equally, whereas AI weights their overlap by the corresponding activation strengths. Global decorrelation therefore need not produce a proportional reduction in active interference.

Together with the retrieval gains in Table~\ref{tab5}, these findings support the complementary roles of the two regularizers: $\mathcal{L}_{\mathrm{coh}}$ constrains shared dictionary geometry, while $\mathcal{L}_{\mathrm{int}}$ addresses the overlap expressed by individual residuals.

\begin{figure}[t]
    \centering
    \includegraphics[width=\columnwidth]
    {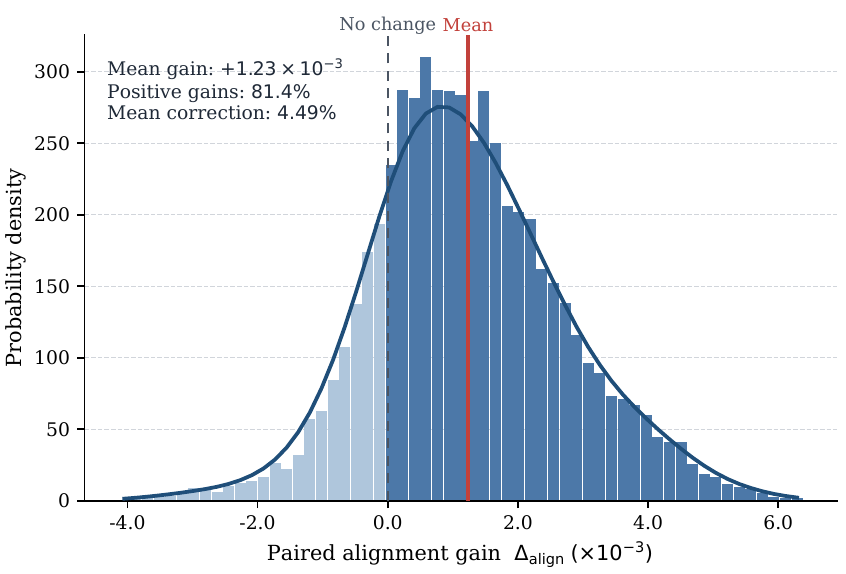}
    \caption{Distribution of paired visual-complement alignment gains on training residuals. Calibration improves alignment for 81.4\% of samples, with a mean gain of $1.23\times10^{-3}$ and a mean relative correction magnitude of 4.49\%.}
    \label{fig4}
\end{figure}

\subsubsection{Paired Analysis of Visual Complement Calibration}

We examine whether calibration refines sparse textual residuals toward the intended visual complement. For each training sample, we compare the raw residual $\mathbf{r}_i$ and its calibrated counterpart $\tilde{\mathbf{r}}_i$ against the same identity-level RGB--IR direction $\mathbf{c}_{y_i}$. The paired alignment gain is defined as $\Delta_i^{\mathrm{align}}= \operatorname{sim}(\tilde{\mathbf{r}}_i,\mathbf{c}_{y_i}) -\operatorname{sim}(\mathbf{r}_i,\mathbf{c}_{y_i})$, where $\operatorname{sim}(\cdot,\cdot)$ denotes cosine similarity. Positive values indicate improved directional agreement with the visual complement. We also measure the relative correction magnitude as $\delta_i= \|\tilde{\mathbf{r}}_i-\mathbf{r}_i\|_2/ (\|\mathbf{r}_i\|_2)$.

Figure~\ref{fig4} shows the paired gain distribution over the valid training residual samples. Calibration improves alignment for 81.4\% of samples, with a mean gain of $1.23\times10^{-3}$. Meanwhile, the mean relative correction magnitude is only 4.49\%, indicating that these improvements are achieved through small adjustments to the original residuals. This behavior is consistent with the residual adapter design: identity-level visual supervision guides the refinement of sample-specific textual evidence while retaining the sparse residual as the basis for compensation.

The combination of broadly positive alignment gains and small correction magnitudes supports calibration as a refinement of sparse fusion. The minority of negative gains shows that improvement is not uniform across samples, but does not identify the cause of these exceptions. As a training-set diagnostic, this analysis verifies the adapter's response to visual-complement supervision; the retrieval experiments separately assess its utility on unseen identities.

\begin{table*}[t]
\centering
\scriptsize
\caption{Robustness analysis under textual corruption. Token-level masking and identity-mismatched descriptions are evaluated to examine the robustness of SARC against incomplete and unreliable textual guidance. Rank-1 (\%), mAP (\%), and mINP (\%) are reported.}
\label{tab10}
\setlength{\tabcolsep}{5pt}

\begin{tabular}{lcccccccccccc}
\toprule
\multirow{2}{*}{Text Condition}
& \multicolumn{6}{c}{Tri-SYSU-MM01}
& \multicolumn{6}{c}{Tri-LLCM} \\
\cmidrule(lr){2-7}
\cmidrule(lr){8-13}

& \multicolumn{3}{c}{TVI-LFM}
& \multicolumn{3}{c}{SARC}
& \multicolumn{3}{c}{TVI-LFM}
& \multicolumn{3}{c}{SARC} \\

\cmidrule(lr){2-4}
\cmidrule(lr){5-7}
\cmidrule(lr){8-10}
\cmidrule(lr){11-13}

& Rank-1 & mAP & mINP
& Rank-1 & mAP & mINP
& Rank-1 & mAP & mINP
& Rank-1 & mAP & mINP \\
\midrule

Clean
& \textbf{84.90} & 81.47 & 70.85
& 84.80 & \textbf{81.98} & \textbf{71.92}
& 58.19 & 65.08 & 61.83
& \textbf{59.57} & \textbf{66.10} & \textbf{62.99} \\

Token mask 10\%
& 79.59 & 76.98 & 66.04
& \textbf{81.65} & \textbf{78.99} & \textbf{68.36}
& 56.08 & 62.99 & 59.75
& \textbf{57.83} & \textbf{64.40} & \textbf{61.20} \\

Token mask 30\%
& 71.23 & 69.13 & 57.07
& \textbf{75.42} & \textbf{73.20} & \textbf{61.62}
& 52.70 & 59.85 & 56.62
& \textbf{54.96} & \textbf{61.87} & \textbf{58.70} \\

Token mask 50\%
& 63.88 & 62.05 & 49.15
& \textbf{69.29} & \textbf{67.18} & \textbf{54.66}
& 49.75 & 57.06 & 53.80
& \textbf{52.41} & \textbf{59.45} & \textbf{56.24} \\

Identity mismatch
& 21.63 & 25.34 & 17.16
& \textbf{49.76} & \textbf{49.36} & \textbf{36.43}
& 34.18 & 41.28 & 37.84
& \textbf{43.83} & \textbf{50.76} & \textbf{47.17} \\

\bottomrule
\end{tabular}
\end{table*}

\subsection{Robustness Analysis under Imperfect Text Descriptions}

Although text-enhanced VI-ReID methods benefit from auxiliary semantic information, their effectiveness may depend on the quality of the available textual descriptions. In particular, heterogeneous-text residual fusion assumes that the difference between RGB-oriented and IR-oriented descriptions contains reliable identity-related visual cues. However, incomplete or incorrect descriptions may introduce irrelevant semantic variations and disturb the residual construction process. We therefore evaluate the robustness of SARC under imperfect RGB-oriented descriptions by applying token-level text corruption.

Specifically, we randomly mask different proportions of tokens in RGB-oriented descriptions while keeping the IR-oriented descriptions unchanged. The masking ratios are set to 10\%, 30\%, and 50\%. In addition, we consider an identity-mismatched setting, where the RGB-oriented description is randomly replaced by a description belonging to another identity. This setting simulates an extreme case in which the auxiliary description provides semantically plausible but identity-irrelevant information. Results on Tri-SYSU-MM01 and Tri-LLCM are reported in Table~\ref{tab10}.

Under clean descriptions, SARC achieves comparable or superior performance to TVI-LFM across both benchmarks, supporting the utility of residual selection and calibration with reliable textual information. More importantly, SARC exhibits substantially stronger robustness as the textual quality deteriorates. On Tri-SYSU-MM01, when 30\% of RGB-oriented tokens are masked, TVI-LFM decreases from 81.47\% to 69.13\% mAP, whereas SARC maintains 73.20\% mAP. With 50\% token masking, the performance gap further increases, where SARC achieves 67.18\% mAP compared with 62.05\% from TVI-LFM. A similar trend is observed on Tri-LLCM: under 50\% token masking, SARC obtains 59.45\% mAP and 56.24\% mINP, outperforming TVI-LFM by 2.39 and 2.44 percentage points, respectively.

The identity-mismatched setting provides a more challenging evaluation of whether a method can distinguish useful textual variations from incorrect semantic transfer. Replacing the RGB-oriented description with another identity substantially degrades TVI-LFM performance. On Tri-SYSU-MM01, TVI-LFM drops to 25.34\% mAP, whereas SARC retains 49.36\% mAP. Similar robustness is observed on Tri-LLCM, where SARC achieves 50.76\% mAP compared with 41.28\% for TVI-LFM. These results indicate that SARC is less sensitive to identity-inconsistent textual signals. However, its mAP under identity mismatch remains below the image-only backbone (49.36\% vs. 65.84\% on Tri-SYSU-MM01 and 50.76\% vs. 59.41\% on Tri-LLCM), so the method does not eliminate the risk of harmful textual compensation.

The observed robustness is consistent with the design principle of SARC, although these comparisons do not isolate the contribution of each component under corruption. Unlike direct residual fusion, which assumes that all textual differences are equally transferable, SARC first selects sparse feature-strength variations and suppresses overlapping directions through low-interference learning. Furthermore, visual complement calibration provides identity-level supervision that encourages the residual to encode appearance changes consistent with the target identity. These mechanisms may reduce the influence of unreliable textual variations on the final residual representation.

Overall, these experiments demonstrate that SARC does not merely improve retrieval accuracy under ideal textual conditions, but also provides a more stable residual construction mechanism when auxiliary descriptions are incomplete or identity-mismatched. This supports the central motivation of SARC that textual differences should be selectively validated rather than directly regarded as visual complements.

\begin{table}[t]
\centering
\scriptsize
\caption{Efficiency comparison between TVI-LFM and SARC under identical settings on one NVIDIA RTX 4090 GPU. Time is per-query latency at batch size 1. FLOPs follow the MAC convention, counting one multiply--accumulate as one operation, for the query-processing scope described in the text.}
\label{tab11}
\setlength{\tabcolsep}{8pt}
\begin{tabular}{lccc}
\toprule
Method & Params (M) & FLOPs (G) & Latency (ms/query) \\
\midrule
TVI-LFM 
& 122.64 
& 22.50 
& 18.40 \\

SARC 
& 156.75 
& 22.60 
& 19.02 \\

\bottomrule
\end{tabular}
\end{table}

\subsection{Efficiency Analysis}

We evaluate the computational efficiency of SARC in terms of parameter size, computational cost, and inference latency. As reported in Table~\ref{tab11}, SARC introduces additional parameters mainly from the sparse residual decomposition module, including the sparse encoder and dictionary used to model textual residuals. Although SARC increases the total parameters from 122.64M to 156.75M, corresponding to an approximately 27.81\% increase, most of these parameters are involved in residual representation learning rather than high-resolution visual feature extraction.

Despite the increased parameter count, SARC adds relatively few FLOPs in the reported comparison. Specifically, the inference cost increases from 22.50G to 22.60G FLOPs, with an approximately 0.44\% increase over the TVI-LFM baseline. This is because the proposed calibration process operates on compact feature representations and does not introduce additional image-level feature extraction or expensive cross-modal interaction modules.

Furthermore, we measure the practical inference latency under the same hardware and evaluation protocol using a single NVIDIA RTX 4090 GPU with FP32 precision. Each query contains one $288\times144$ IR image and two 77-token text sequences (the full RGB-oriented description and the filtered IR-oriented description). Timing includes visual encoding, both complete text-encoding passes, residual fusion, and the deployed output heads. At batch size 1, we perform 20 warm-up iterations and measure the average per-query latency over repeated iterations using CUDA events, with GPU synchronization before and after the timed loop. SARC increases the per-query latency from 18.40 ms/query to 19.02 ms/query, resulting in an approximately 3.37\% overhead based on the displayed values. These results demonstrate that SARC achieves effective textual residual calibration with limited additional computational burden, within the reported text-assisted inference setting.

\section{Discussion}
\label{sec6}

The ablation studies and mechanism analyses support treating textual compensation as a process of selection and calibration. Feature-strength decomposition makes textual variations individually accessible, and Top-$k$ coding concentrates them into a compact set of candidate contributions. Low-interference learning reduces dictionary overlap and active residual interference, while visual calibration further improves agreement with identity-level visual complements through small residual corrections. Together, these findings suggest that effective compensation depends on both how textual differences are represented and how their contributions are guided by visual evidence. The complementarity metrics further show that these improvements extend beyond aggregate retrieval accuracy to alignment, discriminative gain, and gain coverage.

A limitation is the reliance on identity-associated RGB-oriented descriptions under the heterogeneous-text protocol. Although the calibration module requires no identity labels or prototypes at inference, this auxiliary description may be unavailable in unconstrained deployments. Future work will investigate obtaining complementary textual cues from independently supplied descriptions or query-based semantic prediction, reducing dependence on identity-associated RGB text while preserving selective and visually calibrated residual fusion.

\section{Conclusion}
\label{sec7}

This work revisits the assumption that textual residuals are inherently visual complements in text-enhanced visible--infrared person re-identification. Under semantic superposition, direct textual subtraction may mix useful appearance cues with redundant linguistic and contextual information. We introduced SARC to address this challenge through selective residual construction and visual calibration. SARC expresses textual variations as sparse feature strengths, suppresses overlap at the dictionary and activation levels, and refines the reconstructed residual using identity-level visual complements and counterfactual supervision. These visual references are required only during training; the learned adapter calibrates query residuals at inference. Experiments and mechanism analyses support the complementary roles of selection, interference suppression, and calibration in improving residual utility and cross-modal retrieval. Our findings suggest that visual complementarity should be explicitly learned and evaluated, rather than presumed from textual differences alone.

\bibliographystyle{main}
\bibliography{main}

@inproceedings{wang2019learning,
  title={Learning to reduce dual-level discrepancy for infrared-visible person re-identification},
  author={Wang, Zhixiang and Wang, Zheng and Zheng, Yinqiang and Chuang, Yung-Yu and Satoh, Shin'ichi},
  booktitle={Proceedings of the IEEE/CVF conference on computer vision and pattern recognition},
  pages={618--626},
  year={2019}
}

@inproceedings{choi2020hi,
  title={Hi-CMD: Hierarchical cross-modality disentanglement for visible-infrared person re-identification},
  author={Choi, Seokeon and Lee, Sumin and Kim, Youngeun and Kim, Taekyung and Kim, Changick},
  booktitle={2020 IEEE/CVF Conference on Computer Vision and Pattern Recognition (CVPR)},
  pages={10254--10263},
  year={2020},
  organization={IEEE}
}

@inproceedings{park2021learning,
  title={Learning by aligning: Visible-infrared person re-identification using cross-modal correspondences},
  author={Park, Hyunjong and Lee, Sanghoon and Lee, Junghyup and Ham, Bumsub},
  booktitle={2021 IEEE/CVF International Conference on Computer Vision (ICCV)},
  pages={12026--12035},
  year={2021},
  organization={IEEE}
}

@inproceedings{wu2021discover,
  title={Discover cross-modality nuances for visible-infrared person re-identification},
  author={Wu, Qiong and Dai, Pingyang and Chen, Jie and Lin, Chia-Wen and Wu, Yongjian and Huang, Feiyue and Zhong, Bineng and Ji, Rongrong},
  booktitle={2021 IEEE/CVF Conference on Computer Vision and Pattern Recognition (CVPR)},
  pages={4328--4337},
  year={2021},
  organization={IEEE}
}

@inproceedings{zhang2022fmcnet,
  title={Fmcnet: Feature-level modality compensation for visible-infrared person re-identification},
  author={Zhang, Qiang and Lai, Changzhou and Liu, Jianan and Huang, Nianchang and Han, Jungong},
  booktitle={2022 IEEE/CVF Conference on Computer Vision and Pattern Recognition (CVPR)},
  pages={7339--7348},
  year={2022},
  organization={IEEE}
}

@article{elhage2022toy,
  title={Toy models of superposition},
  author={Elhage, Nelson and Hume, Tristan and Olsson, Catherine and Schiefer, Nicholas and Henighan, Tom and Kravec, Shauna and Hatfield-Dodds, Zac and Lasenby, Robert and Drain, Dawn and Chen, Carol and others},
  journal={arXiv preprint arXiv:2209.10652},
  year={2022}
}

@inproceedings{huben2024sparse,
  title={Sparse autoencoders find highly interpretable features in language models},
  author={Huben, Robert and Cunningham, Hoagy and Smith, Logan and Ewart, Aidan and Sharkey, Lee},
  booktitle={International Conference on Learning Representations},
  volume={2024},
  pages={7827--7845},
  year={2024}
}

@inproceedings{gao2025scaling,
  title={Scaling and evaluating sparse autoencoders},
  author={Gao, Leo and Dupre la Tour, Tom and Tillman, Henk and Goh, Gabriel and Troll, Rajan and Radford, Alec and Sutskever, Ilya and Leike, Jan and Wu, Jeffrey},
  booktitle={International Conference on Learning Representations},
  volume={2025},
  pages={26721--26754},
  year={2025}
}

@article{dong2024embedding,
  title={Embedding and enriching explicit semantics for visible-infrared person re-identification},
  author={Dong, Neng and Yan, Shuanglin and Zhang, Liyan and Tang, Jinhui},
  journal={arXiv preprint arXiv:2412.08406},
  year={2024}
}

@inproceedings{wu2017rgb,
  title={RGB-infrared cross-modality person re-identification},
  author={Wu, Ancong and Zheng, Wei-Shi and Yu, Hong-Xing and Gong, Shaogang and Lai, Jianhuang},
  booktitle={2017 IEEE international conference on computer vision (ICCV)},
  pages={5390--5399},
  year={2017},
  organization={IEEE}
}

@article{nguyen2017person,
  title={Person recognition system based on a combination of body images from visible light and thermal cameras},
  author={Nguyen, Dat Tien and Hong, Hyung Gil and Kim, Ki Wan and Park, Kang Ryoung},
  journal={Sensors},
  volume={17},
  number={3},
  pages={605},
  year={2017},
  publisher={MDPI}
}

@inproceedings{zhang2023diverse,
  title={Diverse embedding expansion network and low-light cross-modality benchmark for visible-infrared person re-identification},
  author={Zhang, Yukang and Wang, Hanzi},
  booktitle={2023 IEEE/CVF Conference on Computer Vision and Pattern Recognition (CVPR)},
  pages={2153--2162},
  year={2023},
  organization={IEEE}
}

@article{ye2021deep,
  title={Deep learning for person re-identification: A survey and outlook},
  author={Ye, Mang and Shen, Jianbing and Lin, Gaojie and Xiang, Tao and Shao, Ling and Hoi, Steven CH},
  journal={IEEE transactions on pattern analysis and machine intelligence},
  volume={44},
  number={6},
  pages={2872--2893},
  year={2021},
  publisher={IEEE}
}

@inproceedings{ye2020dynamic,
  title={Dynamic dual-attentive aggregation learning for visible-infrared person re-identification},
  author={Ye, Mang and Shen, Jianbing and Crandall, David J. and Shao, Ling and Luo, Jiebo},
  booktitle={European conference on computer vision},
  pages={229--247},
  year={2020},
  organization={Springer}
}

@article{russakovsky2015imagenet,
  title={Imagenet large scale visual recognition challenge},
  author={Russakovsky, Olga and Deng, Jia and Su, Hao and Krause, Jonathan and Satheesh, Sanjeev and Ma, Sean and Huang, Zhiheng and Karpathy, Andrej and Khosla, Aditya and Bernstein, Michael and others},
  journal={International journal of computer vision},
  volume={115},
  number={3},
  pages={211--252},
  year={2015},
  publisher={Springer}
}

@inproceedings{radford2021learning,
  title={Learning transferable visual models from natural language supervision},
  author={Radford, Alec and Kim, Jong Wook and Hallacy, Chris and Ramesh, Aditya and Goh, Gabriel and Agarwal, Sandhini and Sastry, Girish and Askell, Amanda and Mishkin, Pamela and Clark, Jack and others},
  booktitle={International conference on machine learning},
  pages={8748--8763},
  year={2021},
  organization={PmLR}
}

@inproceedings{ye2018hierarchical,
  title={Hierarchical discriminative learning for visible thermal person re-identification},
  author={Ye, Mang and Lan, Xiangyuan and Li, Jiawei and Yuen, Pong},
  booktitle={Proceedings of the AAAI conference on artificial intelligence},
  volume={32},
  number={1},
  year={2018}
}

@inproceedings{dai2018cross,
  title={Cross-modality person re-identification with generative adversarial training},
  author={Dai, Pingyang and Ji, Rongrong and Wang, Haibin and Wu, Qiong and Huang, Yuyu},
  booktitle={Proceedings of the Twenty-Seventh International Joint Conference on Artificial Intelligence},
  pages={677--683},
  doi={10.24963/ijcai.2018/94},
  year={2018}
}

@inproceedings{wang2019rgb,
  title={RGB-infrared cross-modality person re-identification via joint pixel and feature alignment},
  author={Wang, Guan'an and Zhang, Tianzhu and Cheng, Jian and Liu, Si and Yang, Yang and Hou, Zengguang},
  booktitle={2019 IEEE/CVF International Conference on Computer Vision (ICCV)},
  pages={3622--3631},
  year={2019},
  organization={IEEE}
}

@inproceedings{fu2021cm,
  title={CM-NAS: Cross-modality neural architecture search for visible-infrared person re-identification},
  author={Fu, Chaoyou and Hu, Yibo and Wu, Xiang and Shi, Hailin and Mei, Tao and He, Ran},
  booktitle={2021 IEEE/CVF International Conference on Computer Vision (ICCV)},
  pages={11803--11812},
  year={2021},
  organization={IEEE}
}

@inproceedings{yang2022learning,
  title={Learning with twin noisy labels for visible-infrared person re-identification},
  author={Yang, Mouxing and Huang, Zhenyu and Hu, Peng and Li, Taihao and Lv, Jiancheng and Peng, Xi},
  booktitle={2022 IEEE/CVF Conference on Computer Vision and Pattern Recognition (CVPR)},
  pages={14288--14297},
  year={2022},
  organization={IEEE}
}

@inproceedings{ye2021channel,
  title={Channel augmented joint learning for visible-infrared recognition},
  author={Ye, Mang and Ruan, Weijian and Du, Bo and Shou, Mike Zheng},
  booktitle={2021 IEEE/CVF International Conference on Computer Vision (ICCV)},
  pages={13547--13556},
  year={2021},
  organization={IEEE}
}

@inproceedings{fang2023visible,
  title={Visible-infrared person re-identification via semantic alignment and affinity inference},
  author={Fang, Xingye and Yang, Yang and Fu, Ying},
  booktitle={2023 IEEE/CVF International Conference on Computer Vision (ICCV)},
  pages={11236--11245},
  year={2023},
  organization={IEEE}
}

@inproceedings{zhang2022modality,
  title={Modality synergy complement learning with cascaded aggregation for visible-infrared person re-identification},
  author={Zhang, Yiyuan and Zhao, Sanyuan and Kang, Yuhao and Shen, Jianbing},
  booktitle={European conference on computer vision},
  pages={462--479},
  year={2022},
  organization={Springer}
}

@inproceedings{feng2023shape,
  title={Shape-erased feature learning for visible-infrared person re-identification},
  author={Feng, Jiawei and Wu, Ancong and Zheng, Wei-Shi},
  booktitle={2023 IEEE/CVF Conference on Computer Vision and Pattern Recognition (CVPR)},
  pages={22752--22761},
  year={2023},
  organization={IEEE}
}

@inproceedings{kim2023partmix,
  title={Partmix: Regularization strategy to learn part discovery for visible-infrared person re-identification},
  author={Kim, Minsu and Kim, Seungryong and Park, Jungin and Park, Seongheon and Sohn, Kwanghoon},
  booktitle={2023 IEEE/CVF Conference on Computer Vision and Pattern Recognition (CVPR)},
  pages={18621--18632},
  year={2023},
  organization={IEEE}
}

@inproceedings{zhang2021towards,
  title={Towards a unified middle modality learning for visible-infrared person re-identification},
  author={Zhang, Yukang and Yan, Yan and Lu, Yang and Wang, Hanzi},
  booktitle={Proceedings of the 29th ACM international conference on multimedia},
  pages={788--796},
  year={2021}
}

@article{du2024yyds,
  title={Yyds: Visible-infrared person re-identification with coarse descriptions},
  author={Du, Yunhao and Zhao, Zhicheng and Su, Fei},
  journal={arXiv preprint arXiv:2403.04183},
  year={2024}
}

@article{hu2024empowering,
  title={Empowering visible-infrared person re-identification with large foundation models},
  author={Hu, Zhangyi and Yang, Bin and Ye, Mang},
  journal={Advances in Neural Information Processing Systems},
  volume={37},
  pages={117363--117387},
  year={2024}
}

@article{qiu2025advancing,
  title={Advancing visible-infrared person re-identification: Synergizing visual-textual reasoning and cross-modal feature alignment},
  author={Qiu, Yuxuan and Wang, Liyang and Song, Wei and Liu, Jiawei and Shi, Zhiping and Jiang, Na},
  journal={IEEE Transactions on Information Forensics and Security},
  volume={20},
  doi={10.1109/TIFS.2025.3539946},
  pages={2184--2196},
  year={2025},
  publisher={IEEE}
}

@article{lu2026clip,
  title={CLIP-driven dual-level semantic alignment networks for visible-infrared person re-identification},
  author={Lu, Hu and Huang, Hongjin and Wu, Shengli},
  journal={IEEE MultiMedia},
  year={2026},
  publisher={IEEE}
}

@article{huang2026clip,
  title={CLIP-guided progressive body-part semantic alignment for visible-infrared person re-identification},
  author={Huang, Hongjin and Geng, Xia},
  journal={Algorithms},
  volume={19},
  number={7},
  pages={543},
  year={2026},
  publisher={MDPI}
}

@article{xiong2026clip,
  title={CLIP-powered modality centering with spiral training for visible-infrared person re-identification},
  author={Xiong, Jianghao and Xie, Xiaohua and Feng, Qinyu and Lai, Jian-Huang},
  journal={Pattern Recognition},
  volume={177},
  pages={113333},
  doi={10.1016/j.patcog.2026.113333},
  year={2026},
  publisher={Elsevier}
}

@article{dong2025diverse,
  title={Diverse semantics-guided feature alignment and decoupling for visible-infrared person re-identification},
  author={Dong, Neng and Yan, Shuanglin and Zhang, Liyan and Tang, Jinhui},
  journal={IEEE Transactions on Information Forensics and Security},
  volume={20},
  pages={12245--12259},
  year={2025},
  publisher={IEEE}
}

@inproceedings{dong2025richer,
  title={Richer semantics, better alignment: Aligning visual features with explicit and enriched semantics for visible-infrared person re-identification},
  author={Dong, Neng and Yan, Shuanglin and Zhang, Liyan and Tang, Jinhui},
  booktitle={IJCAI},
  pages={927--935},
  year={2025}
}

@article{luo2026visible,
  title={Visible-guided multigranularity prompt learning for visible-infrared person re-identification},
  author={Luo, Yangyan and Chen, Ying},
  journal={Expert Systems with Applications},
  volume={312},
  pages={131464},
  doi={10.1016/j.eswa.2026.131464},
  year={2026},
  publisher={Elsevier}
}

@inproceedings{liu2026visible,
  title={Visible-infrared person re-identification via CLIP-based semantic information fusion},
  author={Liu, Yunqiang and Wu, Jinghua and Zhao, Jianghai and Shi, Xu and Sun, Jincan and Wang, Zihao},
  booktitle={2026 11th International Conference on Intelligent Computing and Signal Processing (ICSP)},
  pages={1794--1801},
  year={2026},
  organization={IEEE}
}

@inproceedings{zhang2025visible,
  title={Visible-infrared person re-identification via mutual reinforcement of prompts and image encoders},
  author={Zhang, Hongde and Ma, Bingpeng},
  booktitle={Proceedings of the 2025 International Conference on Multimedia Retrieval},
  pages={2118--2122},
  year={2025}
}

@article{yu2025clip,
  title={Clip-driven semantic discovery network for visible-infrared person re-identification},
  author={Yu, Xiaoyan and Dong, Neng and Zhu, Liehuang and Peng, Hao and Tao, Dapeng},
  journal={IEEE Transactions on Multimedia},
  volume={27},
  pages={4137--4150},
  year={2025},
  publisher={IEEE}
}

@article{liu2020parameter,
  title={Parameter sharing exploration and hetero-center triplet loss for visible-thermal person re-identification},
  author={Liu, Haijun and Tan, Xiaoheng and Zhou, Xichuan},
  journal={IEEE Transactions on Multimedia},
  volume={23},
  pages={4414--4425},
  year={2020},
  publisher={IEEE}
}

@article{yan2025shapespeak,
  title={ShapeSpeak: Body shape-aware textual alignment for visible-infrared person re-identification},
  author={Yan, Shuanglin and Dong, Neng and Li, Shuang and Yan, Rui and Tang, Hao and Qin, Jing},
  journal={arXiv preprint arXiv:2504.18025},
  year={2025}
}

@article{bricken2023towards,
  title={Towards monosemanticity: Decomposing language models with dictionary learning},
  author={Bricken, Trenton and Templeton, Adly and Batson, Joshua and Chen, Brian and Jermyn, Adam and Conerly, Tom and Turner, Nick and Anil, Cem and Denison, Carson and Askell, Amanda and Lasenby, Robert and Wu, Yifan and Kravec, Shauna and Schiefer, Nicholas and Maxwell, Tim and Joseph, Nicholas and Hatfield-Dodds, Zac and Tamkin, Alex and Nguyen, Karina and McLean, Brayden and Burke, Josiah E. and Hume, Tristan and Carter, Shan and Henighan, Tom and Olah, Christopher},
  journal={Transformer Circuits Thread},
  year={2023}
}

@article{rajamanoharan2024improving,
  title={Improving sparse decomposition of language model activations with gated sparse autoencoders},
  author={Rajamanoharan, Senthooran and Conmy, Arthur and Smith, Lewis and Lieberum, Tom and Varma, Vikrant and Kramar, Janos and Shah, Rohin and Nanda, Neel},
  journal={Advances in Neural Information Processing Systems},
  volume={37},
  pages={775--818},
  year={2024}
}

@inproceedings{li2022counterfactual,
  title={Counterfactual intervention feature transfer for visible-infrared person re-identification},
  author={Li, Xulin and Lu, Yan and Liu, Bin and Liu, Yating and Yin, Guojun and Chu, Qi and Huang, Jinyang and Zhu, Feng and Zhao, Rui and Yu, Nenghai},
  booktitle={European conference on computer vision},
  pages={381--398},
  year={2022},
  organization={Springer}
}

@article{sun2023counterfactual,
  title={Counterfactual attention alignment for visible-infrared cross-modality person re-identification},
  author={Sun, Zongzhe and Zhao, Feng},
  journal={Pattern recognition letters},
  volume={168},
  pages={79--85},
  year={2023},
  publisher={Elsevier}
}

@article{kubaty2026conceptualizing,
  title={Conceptualizing embeddings: Sparse disentanglement for vision-language models},
  author={Kubaty, Piotr and Marsza{\l}ek, Patryk and Struski, {\L}ukasz and Wr{\'o}bel, Adam and Tabor, Jacek and {\'S}mieja, Marek},
  journal={arXiv preprint arXiv:2605.22679},
  year={2026}
}

@article{zhao2026dual,
  title={Dual-space modality consistency learning for universal cross-modal re-identification},
  author={Zhao, Yujian and Zhao, Yukang and Liu, Hankun and Xu, Haoxuan and Li, Bo and Wan, Hanzi and Niu, Guanglin},
  journal={arXiv preprint arXiv:2608.06943},
  year={2026}
}

@article{li2026mfen,
  title={MFEN: Multi-frequency expert network for visible-infrared person re-ID},
  author={Li, Xulin and Lu, Yan and Liu, Bin and Yang, Qinhong and Chu, Qi and Gong, Tao and Yu, Nenghai},
  journal={arXiv preprint arXiv:2606.12051},
  year={2026}
}

@article{yang2026clip4vi,
  title={CLIP4VI-ReID: Learning modality-shared representations via CLIP semantic bridge for visible-infrared person re-identification},
  author={Yang, Xiaomei and Gao, Xizhan and Niu, Sijie and Zhu, Fa and Feng, Guang and Qu, Xiaofeng and Camacho, David},
  journal={IEEE Transactions on Biometrics, Behavior, and Identity Science},
  year={2026},
  publisher={IEEE}
}

\end{document}